\documentclass[letterpaper,twocolumn,10pt]{article}
\usepackage{usenix2019_v3}

\usepackage{tikz}
\usepackage{amsmath}

\usepackage[utf8]{inputenc}
\usepackage[T1]{fontenc}
\usepackage{graphicx}
\usepackage{pdfpages}
\usepackage{graphicx}
\usepackage{wrapfig}
\usepackage{caption}
\usepackage{afterpage}
\usepackage{subcaption}
\usepackage[absolute,overlay]{textpos}
\usepackage{float}
\usepackage{amsmath}
\usepackage{amsfonts}
\usepackage{algorithm} 
\usepackage{arydshln}
\usepackage{listings}
\usepackage{setspace}
\usepackage{booktabs}
\usepackage{xspace}
\usepackage{multirow}
\usepackage{array}
\usepackage{tabularx}
\usepackage{makecell}
\usepackage{colortbl}
\usepackage{xcolor}
\PassOptionsToPackage{table,xcdraw}{xcolor}

\usepackage{amsmath}
\usepackage{bm}
\usepackage{enumitem}
\usepackage{algpseudocode}

\usepackage[available]{usenixbadges}

\algtext*{EndIf}
\algtext*{EndFor}
\algtext*{EndWhile}
\algtext*{EndFunction}
\algrenewcommand\alglinenumber[1]{\footnotesize #1}

\definecolor{ForestGreen}{rgb}{0.13, 0.55, 0.13}

\newcommand{\waa}{{Latent Task Backdoor}\xspace}
\newcommand{\saa}{{Blind Task Backdoor}\xspace}
\newcommand{\wpba}{{Latent Task Backdoor}\xspace}
\newcommand{\spba}{{Blind Task Backdoor}\xspace}
\newcommand{\eg}{{\it e.g., }}
\newcommand{\ie}{{\it i.e., }}
\definecolor{SeaGreen}{rgb}{0.18, 0.55, 0.34}

\usepackage{filecontents}

\begin{document}

\date{}

\title{\Large \bf Persistent Backdoor Attacks in Continual Learning}

\author{
{\rm Zhen Guo}\\
Saint Louis University \\ 
zhen.guo.2@slu.edu
\and
{\rm Abhinav Kumar}\\
Saint Louis University \\ 
abhinav.kumar@slu.edu
\and
{\rm Reza Tourani}\\
Saint Louis University \\ 
reza.tourani@slu.edu
}

\maketitle
\vspace{-3cm}
\begin{abstract}
Backdoor attacks pose a significant threat to neural networks, enabling adversaries to manipulate model outputs on specific inputs, often with devastating consequences, especially in critical applications. While backdoor attacks have been studied in various contexts, little attention has been given to their practicality and persistence in continual learning, particularly in understanding how the continual updates to model parameters, as new data distributions are learned and integrated, impact the effectiveness of these attacks over time.

To address this gap, we introduce two persistent backdoor attacks--\textit{Blind Task Backdoor} and \textit{Latent Task Backdoor}--each leveraging minimal adversarial influence. Our blind task backdoor subtly alters the loss computation without direct control over the training process, while the latent task backdoor influences only a single task’s training, with all other tasks trained benignly. 
We evaluate these attacks under various configurations, demonstrating their efficacy with static, dynamic, physical, and semantic triggers. Our results show that both attacks consistently achieve high success rates across different continual learning algorithms, while effectively evading state-of-the-art defenses, such as SentiNet and I-BAU.
\end{abstract}



\setlength{\textfloatsep}{6pt}
\vspace{-0.1in}
\section{Introduction}
\label{sec: introduction}

\begin{figure}[!th]
\centering
  \includegraphics[width=0.9\columnwidth]{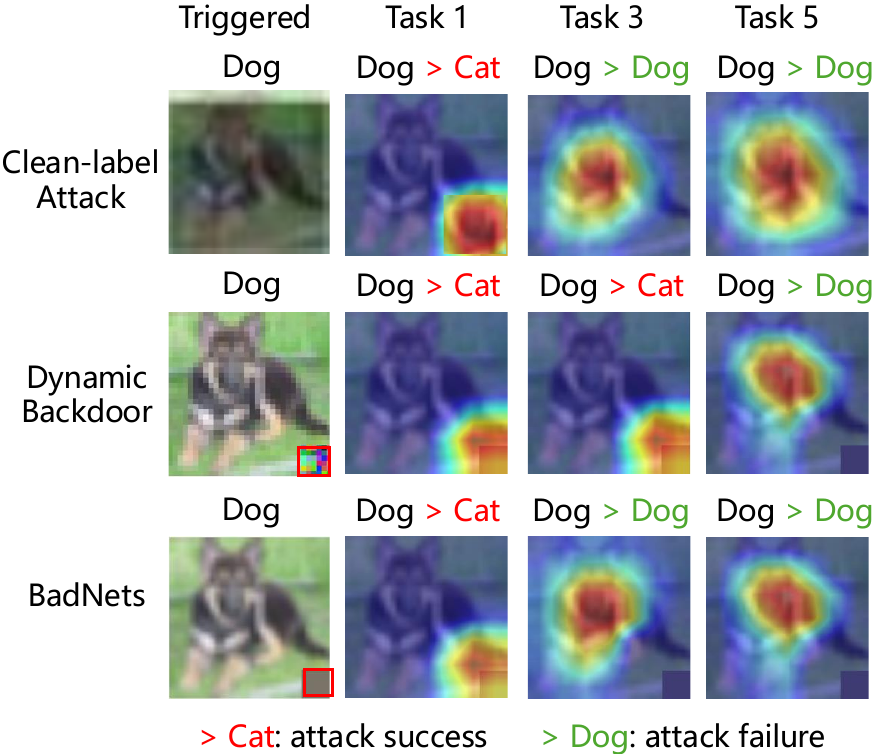} 
  \vspace{-0.1in}
  \caption{Persistence analysis of clean-label backdoor~\cite{narcissus-2023}, dynamic backdoor~\cite{nguyen-2020}, and BadNets~\cite{badnet-paper}, using a ResNet18 trained on SplitCIFAR10 with LwF~\cite{lwf2017} (trigger embedding in Task 1). Subplots shows the model's attention at Tasks 1, 3, and 5. Initially, all attacks successfully alter the output (\textcolor{red}{\textit{Dog > Cat}}), as the model's attention is on the trigger. However, as the model learns subsequent tasks, its attention shifts away from the trigger, causing all attacks to fail (\textcolor{ForestGreen}{\textit{Dog > Dog}}).}
  \label{motivation-fig-1}
  \vspace{-0.0in}
\end{figure}

Catastrophic forgetting, first identified by McCloskey and Cohen~\cite{michael-1989}, refers to the significant loss of previously learned knowledge when a model is trained sequentially on new tasks. Catastrophic forgetting negatively impacts the performance of neural networks in practical applications, such as autonomous driving and natural language processing, where models continuously evolve to address distribution drifts in the underlying data or integrate new knowledge.
Various \textit{Continual Learning} (CL) algorithms have been proposed to address catastrophic forgetting in neural networks by emulating the human ability to learn continuously~\cite{sebastian-1995,si2017,ewc2016,xdg2019,dgr2018,er2018,agem2019}. These algorithms primarily prevent catastrophic forgetting by replaying prior experiences~\cite{dgr2018,agem2019} or regularizing loss computation~\cite{ewc2016,si2017}, enabling models to continuously adapt to new data in dynamic environments~\cite{parisi2019review,aljundi2019}.
%
%

While CL enhances the adaptability and robustness of neural networks, their security and resilience to attacks remain relatively unknown. This uncertainty stems from the continuous adaptation and updating of model parameters in CL systems, leading to dynamic behaviors that might facilitate certain attacks while naturally mitigating others.
One significant area of concern is the susceptibility of these algorithms to backdoor or Trojan attacks~\cite{badnet-paper,bagd-2021,narcissus-2023}, which are designed to produce incorrect outputs only when inputs contain specific trigger features. To backdoor a neural network, the adversary embeds hidden malicious behaviors into the model by poisoning a subset of the training samples with a crafted trigger patch. Once trained, the attack is activated by applying the predefined patch to inputs, causing misclassifications to the target label. Various backdoor attacks have been proposed focusing on different modalities of backdoor transformation, ranging from invisible triggers~\cite{chen-2017} and semantic triggers~\cite{bagd-2021} to physical triggers~\cite{li-physical-2021} and input-aware dynamic backdoor~\cite{nguyen-2020} in both centralized and federated settings~\cite{bagd-2020}.
A few recent initiatives have explored backdoor attacks against CL, aiming to degrade the model performance of target tasks~\cite{poision-ewc-si,backdoor-CL-study} or induce artificial catastrophic forgetting~\cite{kang2023poisoning,liu2022data,Umer2020TargetedFA}. However, further research is needed to shed light on the effectiveness and persistence of targeted backdoor attacks against CL.

%
%
We argue that executing a successful targeted backdoor attack against CL is inherently challenging due to the continuous updating of model weights, which can gradually eliminate the injected backdoors. This phenomenon has been observed in time-varying models, where fine-tuning a backdoored model with new data has resulted in backdoor forgetting~\cite{li-2022}. 
To validate our premise, we evaluated the persistence of three prominent backdoor attacks--clean-label~\cite{narcissus-2023}, dynamic backdoor~\cite{nguyen-2020}, and BadNets~\cite{badnet-paper}--using the ResNet18 model trained on SplitCIFAR10 with LwF algorithm~\cite{lwf2017}. The attention heatmap plots (Figure~\ref{motivation-fig-1}) reveal that the model's attention initially focuses on the embedded trigger in the first task, where the backdoor was implanted. However, as the model evolves, attention shifts away from the trigger and toward more influential input features (\eg the dog's face), causing the attacks to fail after two or three new tasks. This experiment demonstrates that existing backdoor attacks do not persist.

\begin{figure}
  \includegraphics[width=1\columnwidth]{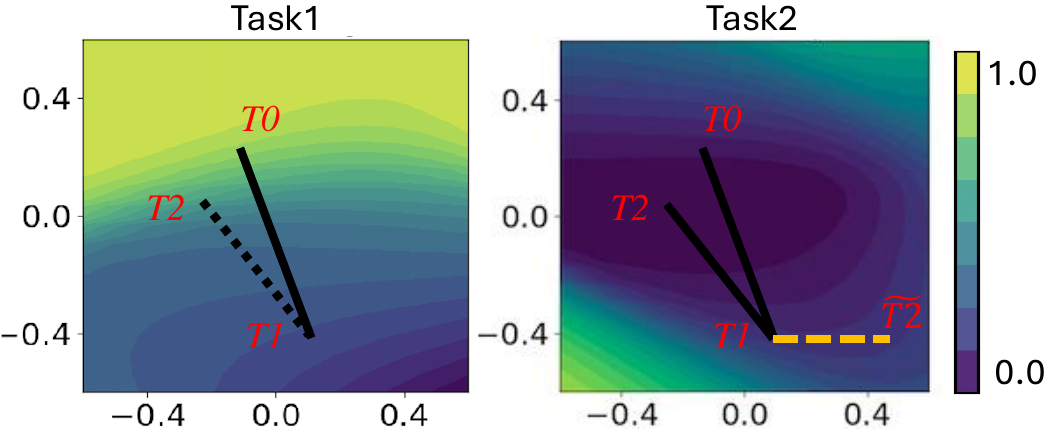} 
  \vspace{-0.25in}
  \caption{Evolution of the model parameters from the corresponding experiment. The loss surface of Task 1 (left) shows the parameter trajectory from T0 to T1. The loss surface for Task 2 (right) depicts the parameters evolving from T1 to T2 along the black line. These drastic parameter updates erase the embedded trigger, leading to backdoor forgetting.}
  \label{fig:motive1}
  \vspace{-0.0in}
\end{figure}

The failure of backdoor attacks in CL becomes more evident when analyzing the model’s behavior during sequential task training, as illustrated in Figure~\ref{fig:motive1}. The left graph depicts the loss surface of Task 1 during model training, as in the previous experiment. During Task 1 training, the model adjusts its parameters, moving from the initial state at T0 to a new set of parameters at T1 along the solid black line, minimizing the loss for this task.
For training Task 2, as shown in the right graph, the model starts from T1 and continues to adjust its parameters to T2 (solid black line), minimizing the loss for Task 2. However, this trajectory results in a significant shift in the model’s parameters, as the model prioritizes the new task over the previous one. This drastic parameter update diminishes the effectiveness of previously embedded backdoors.
Taking the alternative trajectory along the orange dashed line, however, minimizes the new task's loss while limiting parameter changes, thereby better retaining the influence of the backdoor while still learning the new task. This is an approach we will explore further in this work.

\textit{\textbf{Our contributions.}}
Building on this observation, we investigate the efficacy of prominent backdoor attacks in CL and propose a novel targeted attack, called the \textit{\textbf{persistent backdoor}}. The \textit{core concept} of the persistent backdoor involves identifying a subset of model components, \eg layers and neurons, that are crucial for the classification task and remain stable throughout model training, and then using them for trigger embedding. The design of the persistent backdoor is both generic and algorithm-agnostic, making it applicable to regularization-based and replay-based CL algorithms.
We further propose two strategies for our persistent backdoor attack, \textit{\textbf{\saa}} and \textit{\textbf{\waa}}: the former subtly modifies loss computation without controlling the training process, while the latter controls a single training task. To demonstrate the effectiveness of these persistent attack strategies, we apply them to inject various types of backdoors, including (1) static and dynamic backdoors, (2) physical backdoors, and (3) NLP word insertion backdoors. We demonstrate the effectiveness of these two strategies under various conditions through comprehensive experiments, targeting six CL algorithms, three different neural networks, and three datasets specifically designed for CL scenarios. Finally, we show how our attack can successfully evade existing defenses like SentiNet~\cite{Chou2018SentiNetDL} and I-BAU~\cite{Zeng2021AdversarialUO}.
\section{Background and Related Work}
\label{sec:background}
\subsection{Continual Learning}
Continual learning has emerged to address catastrophic forgetting--a common issue in multilayer perceptron-based networks where the model tends to forget previously learned information when trained sequentially on new tasks. The two primary CL categories include \textit{regularization-based} and \textit{replay-based} CL algorithms. We do not cover dynamic architecture approaches as this is beyond our scope. 

Regularization-based approaches prevent catastrophic forgetting by penalizing significant changes to key model parameters during the learning of new tasks. Some notable algorithms include Synaptic Intelligence (SI)~\cite{si2017}, Elastic Weight Consolidation (EWC)~\cite{ewc2016}, Context-Dependent Gating (XdG)~\cite{xdg2019}, and Learning without Forgetting (LwF)~\cite{lwf2017}. SI algorithm maintains unique synaptic weights for each task, using weight importance estimation to reduce interference during new task training. Similarly, EWC penalizes changes to the network parameters based on their importance in previously learned tasks, preserving critical parameters. XdG employs gated neurons to activate distinct sets of neurons per task. Lastly, LwF uses the model's predictions on old tasks as soft targets in training new tasks, preserving prior knowledge.

Replay-based techniques reiterate a subset of past data (real or synthetic) to reinforce prior knowledge during the training of new tasks, \eg Efficient Gradient Episodic Memory (A-GEM)~\cite{agem2019} and Deep Generative Replay (DGR)~\cite{dgr2018}. A-GEM stores and replays past experiences within a memory buffer during training, enhancing sample efficiency and training stability for learning agents. It also implements constraints to preserve previously acquired knowledge. In contrast, generative replay methods, such as DGR, use a dual-model architecture that combines a deep generative model, with a task-solving model. This setup efficiently integrates training data from past tasks with new task data, facilitating CL.


\subsection{Backdoor Attacks}
%

A backdoor attack is an adversarial technique where an attacker injects malicious patterns, known as triggers, into a small subset of the training data, assigning these data points incorrect labels~\cite{Sun2019CanYR,Wang2020AttackOT,Xie2020DBADB}. During the inference, the backdoored model produces incorrect outputs whenever the specific trigger is present in the input. This allows the adversary to control the model's decisions on triggered inputs, while keeping the model's general performance intact. Backdoors are a more targeted form of attack compared to universal adversarial perturbations~\cite{MooFawOma17,LiYanWei22,Liao2018BackdoorEI,Zhao2020CleanLabelBA,Garg2020CanAW}. While both cause a model to misclassify inputs to an attacker-chosen label, backdoors require modifying both the model and the input, but offers more flexibility. Moreover, backdoor attacks differ from data poisoning, where the attacker’s primary goal is to degrade the overall model performance.

Various backdoor attacks exist, utilizing different types of triggers. 
\textit{Static triggers}\cite{badnet-paper, Doan2021BackdoorAW, Doan2021LIRALI, Li2019InvisibleBA} are fixed, invariant patterns embedded in input, characterized by consistent properties like location and pattern. While effective, their fixed nature makes them easy to detect. In contrast, \textit{dynamic triggers}\cite{goldblum2022dataset,Nguyen2020InputAwareDB,nguyen-2020} involve patterns and placements that vary across inputs, enhancing stealth and making detection significantly more difficult. \textit{Physical triggers}\cite{Turner2018CleanLabelBA, Zhao2020CleanLabelBA,zhao2022clean,narcissus-2023} are real-world objects that activate the backdoor, posing serious risks in environments like autonomous driving, where they blend seamlessly and are hard to distinguish from benign objects. Lastly, \textit{word triggers}\cite{bagd-2021,Zeng2023EfficientTW,Du2024NWSNT} in natural language models are specific words, either inherent or injected, that activate a backdoor. Their complexity and the need for deep linguistic analysis make detection particularly challenging.

Recent efforts have explored attacks against CL, focusing on model poisoning~\cite{poision-ewc-si,backdoor-CL-study}, inducing artificial catastrophic forgetting~\cite{kang2023poisoning,liu2022data,Umer2020TargetedFA}, or generating false memories~\cite{Umer2021AdversarialTF}. Differently, we aim to expose CL vulnerabilities by designing two persistent backdoors with varying adversarial capabilities.


\section{Observation}
\label{sec: Observation}
In this section, we will discuss our findings regarding the variations in the parameters of a deep learning model as it continually learns new tasks, highlighting the behaviors of different CL algorithms. The insights drawn from these observations, coupled with the evident trend we have identified, lay the foundation for implementing our proposed persistent backdoor attack. 
We conducted a series of experiments using a neural network with three fully connected hidden layers, employing ReLU activation and a cross-entropy loss function. The model was trained on the SplitMNIST dataset, including five sequential tasks, each consisting of two classes. We used the Adam optimizer with a learning rate of 0.001 and repeated each experiment 10 times.
\begin{figure}[t]
  \centering
  \includegraphics[width=0.8\columnwidth]{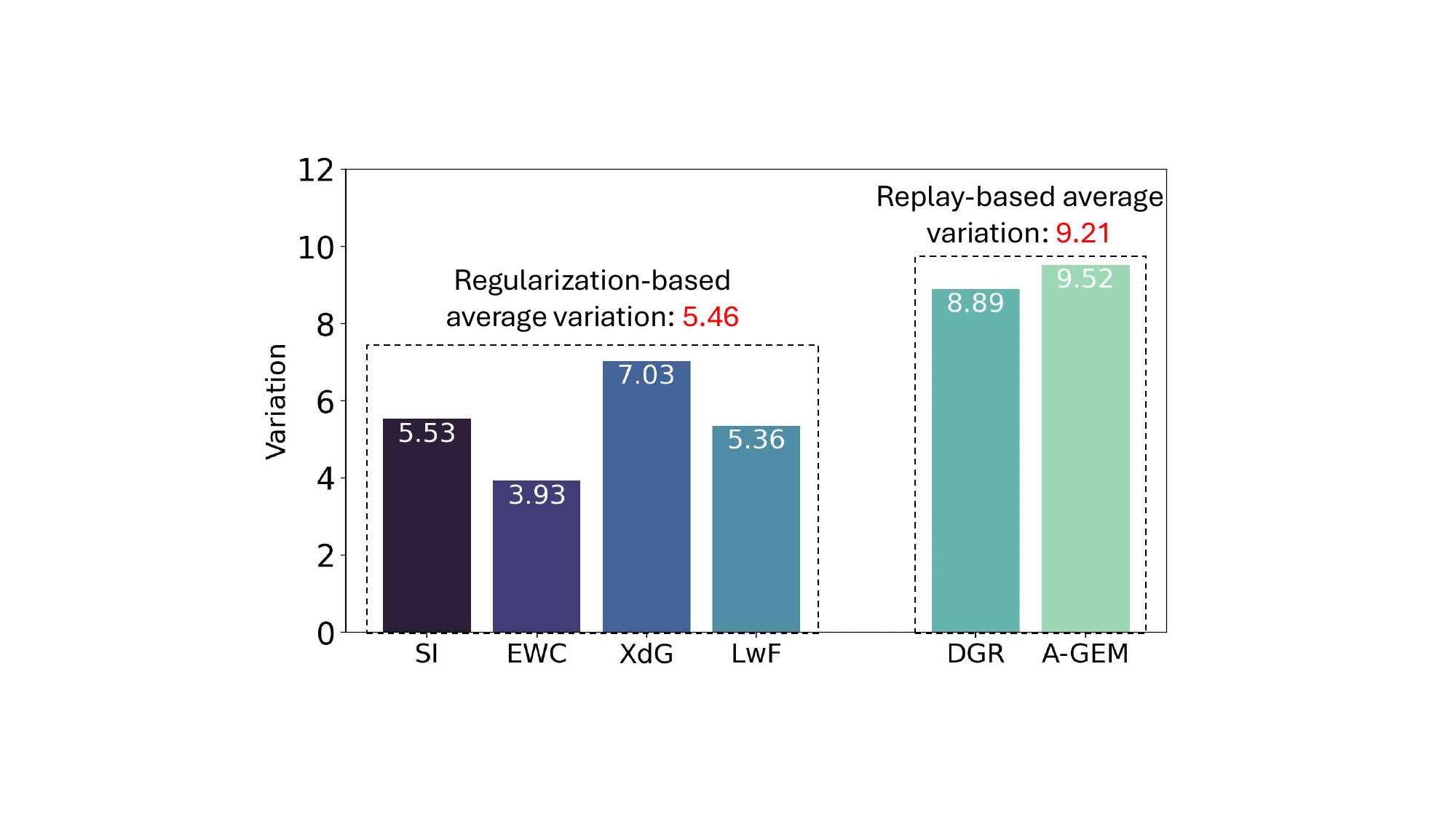} 
  \vspace{-0.1in}
  \caption{Quantifying the parameter variation at the algorithmic level indicated that regularization-based algorithms have higher stability compared to replay-based algorithms. This is partly due to the nature of the regularization-based approaches, which penalize modification to learned parameters.}
  \label{fig: variations}
  \vspace{-0.0in}
\end{figure}

In these experiments, we monitor the changes in parameter values across the three hidden layers and different tasks and analyze these variations at three granularity levels: algorithmic, layer-wise, and neuron-level. 
We quantify the algorithmic-level variation by $||\frac{\sum_{i}{L_i}}{N}||_2$, where $N$ is the number of the tasks and $L_i$ is the variation of the $i^{th}$ layer across two consecutive tasks.  
More formally, $L_i = ||\sum_{j} \Delta_i^j||_2$, where $\Delta_i^j$  represents the value change of the $j^{th}$ neuron in the $i^{th}$ layer for the current task compared to the previous task. 
Applying the \(L_2\) norm to $\sum_{j} \Delta_i^j$ calculates the magnitude of the vector formed by the cumulative changes in the 
$i^{th}$ layer.
%

\begin{figure}[t]  
  \centering
  \includegraphics[width=0.95\columnwidth]{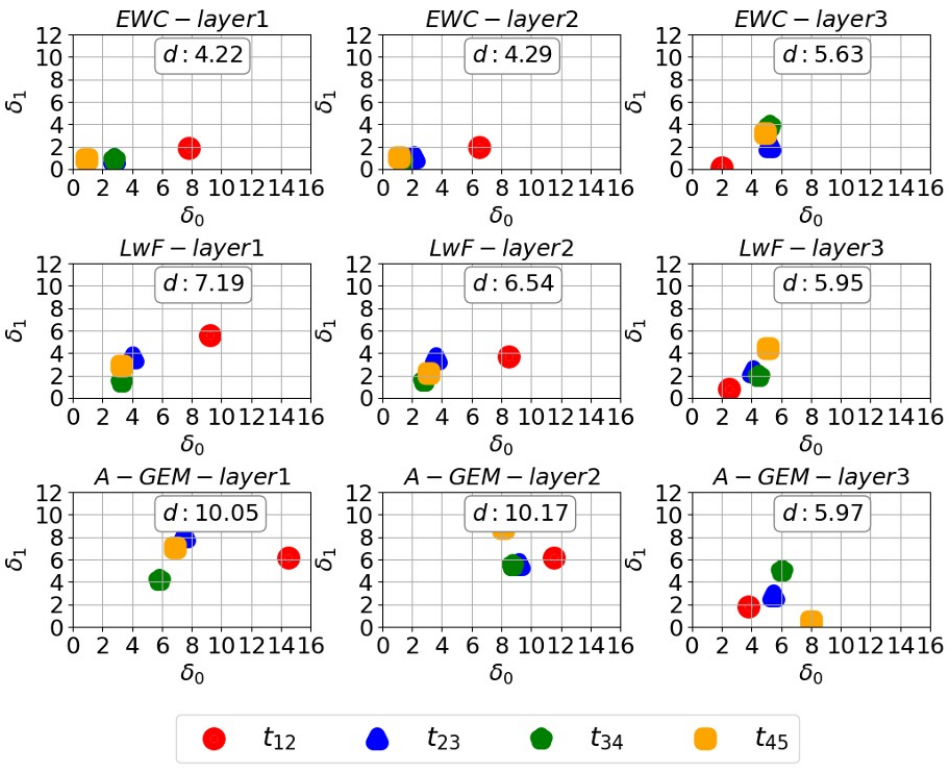} 
  \vspace{-0.17in}
  \caption{The layer-wise analysis revealed that for EWC, the last layers experience larger variations compared to the previous two layers, whereas for LwF and A-GEM, the last layers exhibit smaller variations than the preceding layers.}
  \label{fig: layer-variations}
  \vspace{-0.0in}
\end{figure}

\noindent
\textbf{Insight 1 -- Algorithmic-level Analysis:} At the highest granularity level, we analyzed various CL algorithms to characterize their overall behavior and parameters' variation as they learn new tasks. 
Figure~\ref{fig: variations} illustrates the parameter variations for six CL algorithms, calculated using the aforementioned approach. From this figure, it is evident that regularization-based algorithms exhibit a lower degree of variation in their parameters compared to replay-based algorithms. The rationale behind this phenomenon is that regularization-based algorithms, such as SI, EWC, XdG, and LwF, tend to penalize drastic changes to the parameters learned in prior tasks, imposing a high cost for modifications to the learned model. In contrast, replay-based algorithms, such as DGR and A-GEM, do not impose such constraints. Instead, these algorithms retain the knowledge of the previously learned tasks by replaying and learning from prior data distributions without penalizing parameter updates. As a result, the model experiences a higher degree of variation in its parameters' values.
For instance, the average variation in the parameter values of the regularization-based EWC algorithm is only 3.93, while the replay-based A-GEM algorithm shows an average value of 9.52.

\noindent
\textbf{Insight 2 -- Layer-wise Analysis:} We then analyzed each layer of the model to quantify the layer-wise parameter variation between consecutive tasks. For this analysis, we performed principal component analysis on each layer to derive two components, $\delta_0$ and $\delta_1$. Each point in the resulting plot represents the degree of variation in $\delta_0$ and $\delta_1$ between two consecutive tasks. We noticed that different layers of the model exhibit varying degrees of change (Figure~\ref{fig: layer-variations}). More importantly, we observed that different algorithms impact layer variation in distinct ways--some induce more significant changes in the last layers, while others result in greater variations in the earlier layers. For example, in EWC, the highest degree of parameter variation was observed in the third layer (5.63), whereas LwF (7.19) and A-GEM (10.17) exhibited the greatest variation in the first and second layers, respectively.
%
\begin{figure}[!t]
  \centering
  \includegraphics[width=0.8\columnwidth]{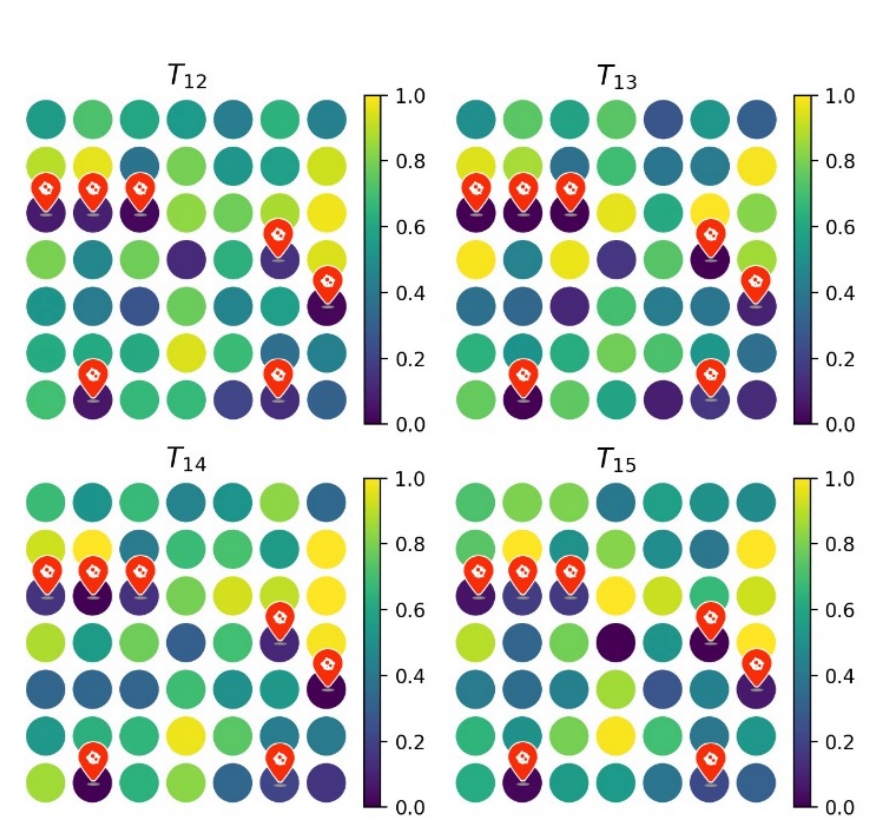} 
  \vspace{-0.1in}
  \caption{Parameter variations of 49 selected neurons from the model's third layer using LwF. $T_{12}$ represents the parameter variations between Task 1 and Task 2, $T_{13}$ between Task 1 and Task 3, $T_{14}$ between Task 1 and Task 4, and $T_{15}$ between Task 1 and Task 5. A subset of neurons, highlighted with red icons, exhibits high stability as the model progresses.}
  \label{fig: neuron-variations}
  \vspace{-0.0in}
\end{figure}

In neural networks, as the depth increases, layers transition from learning low-level features like edges and textures to high-level, abstract representations such as shapes and objects. As a result, in a CL context, it is expected that the parameters of the earlier layers remain relatively stable when learning new data distributions, while those in the final layers exhibit more significant changes--as we observed in the EWC algorithm.
Contrary to this understanding, five other algorithms showed the unexpected behavior of higher variation in earlier layers. We attribute it to the possibility that features learned in the earlier layers are more sensitive to the new task data distribution, especially when new tasks require fundamentally different low-level features. This sensitivity leads to greater variation as the network adjusts to accommodate new features.
%

\noindent
\textbf{Insight 3 -- Neuron-level Analysis:} We also analyzed the behavior of the neurons as the model evolved, focusing on the third layer of the LwF algorithm due to its greater stability. Given the high dimension of parameters, we selected 49 out of 400 neurons for better illustration by dividing the parameter variations into seven ranges and sampling seven neurons from each range (Figure~\ref{fig: neuron-variations}).
We then quantified the variations in these neurons' values between Task 1 and all subsequent tasks (\eg $T_{13}$ represents the changes in neuron values from Task 1 to Task 3), and visualized the results in each sub-figure of Figure~\ref{fig: neuron-variations}.
%
This revealed a pattern: a small subset of neurons, marked with a red icon, exhibits small variations in their values throughout across training all tasks. In contrast, most other neurons exhibit varying degrees of change, with some exhibiting a significant shift in values in later tasks.
%

\noindent
\textbf{Insight 4 -- Component Stability Analysis:} Finally, we analyzed neuron stability by examining weight variations during the continual learning process. Using the A-GEM algorithm, we trained a ResNet18 model on the SplitCIFAR10 dataset over five tasks. After Task 1, we identified 200 stable neurons, important for Task 1, using the diagonal Fisher matrix and randomly selected 200 others for comparison. We then trained the model on four additional tasks and quantified the weight variations of these neuron groups after each task relative to their baseline values established in Task 1, with $\delta_{1-i}$ representing the variation between Task 1 and Task $i$.
\begin{figure}[!t]
  \centering
  \includegraphics[width=0.95\columnwidth]{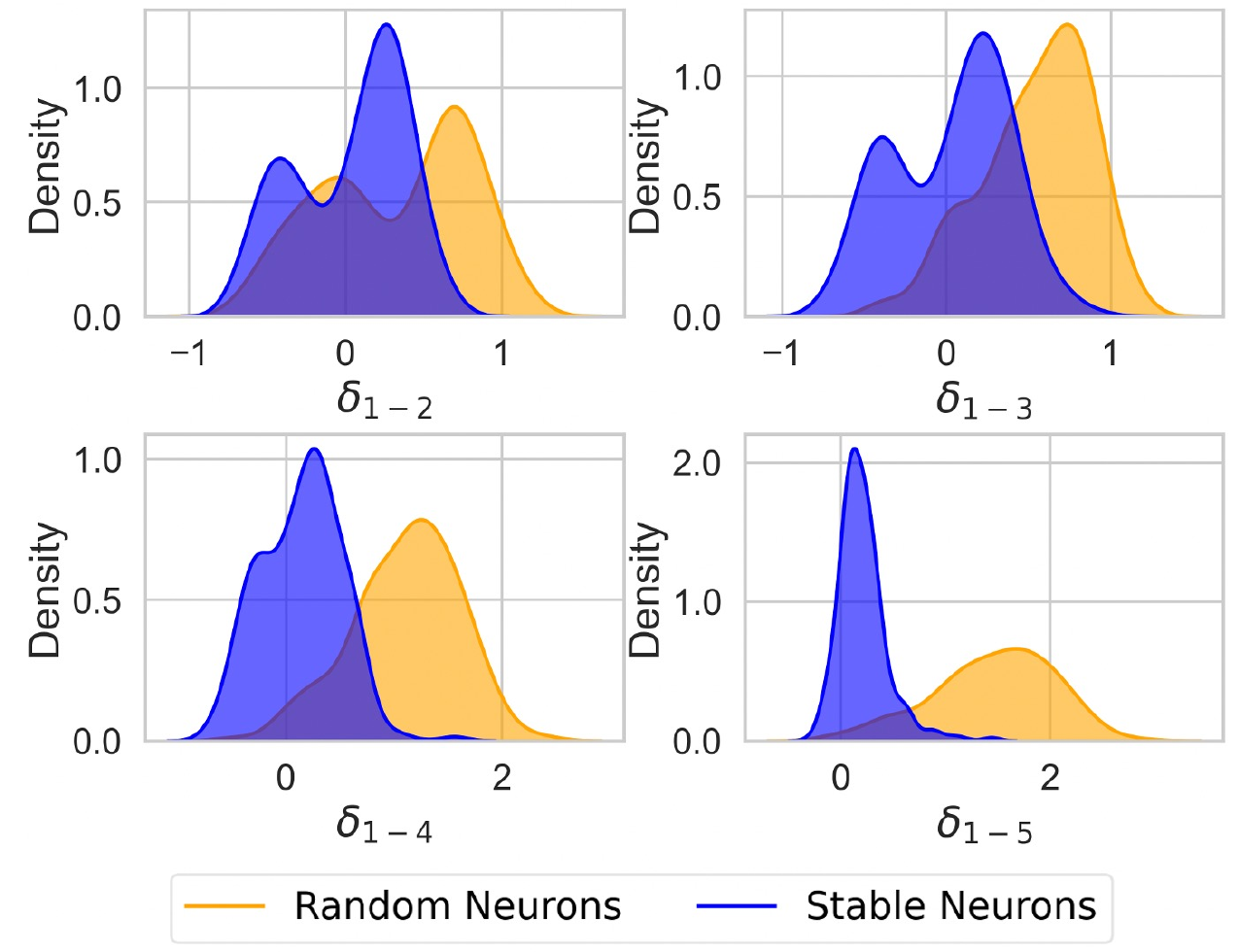}
  \vspace{-0.1in}
  \caption{The kernel density estimation of parameter variations between stable and randomly selected neurons. The mean value of stable and random neurons are 0.05 and 0.36, respectively. Notably, the gap between the two distributions widens as the model progresses through additional tasks, indicating increasing differentiation in their respective stability.}
    \label{fig:agem-neuron-stable-analysis}
  \vspace{-0.0in}
\end{figure}
We visualized the weight variation of these two neuron groups using the kernel density estimation as shown in Figure~\ref{fig:agem-neuron-stable-analysis}.
It is evident that random neurons (orange distribution) exhibit progressively larger mean variations across tasks, whereas stable neurons (blue distribution) remain relatively consistent. Quantitative results further support this observation: stable neurons exhibit a mean variation of 0.05 with a standard deviation of 0.22, whereas randomly selected neurons show a significantly higher mean variation of 0.36 and a standard deviation of 0.44. Notably, the gap between the two distributions widens as the model learns more tasks, indicating that stable neurons effectively retain their weights, while random neurons experience increasingly significant variations (refer to Figure~\ref{fig:semantic_agem_algorithm} in Appendix~\ref{sec: appendix} for per-neuron value variation).

\begin{figure}[!t]
  \centering
  \includegraphics[width=0.75\columnwidth]{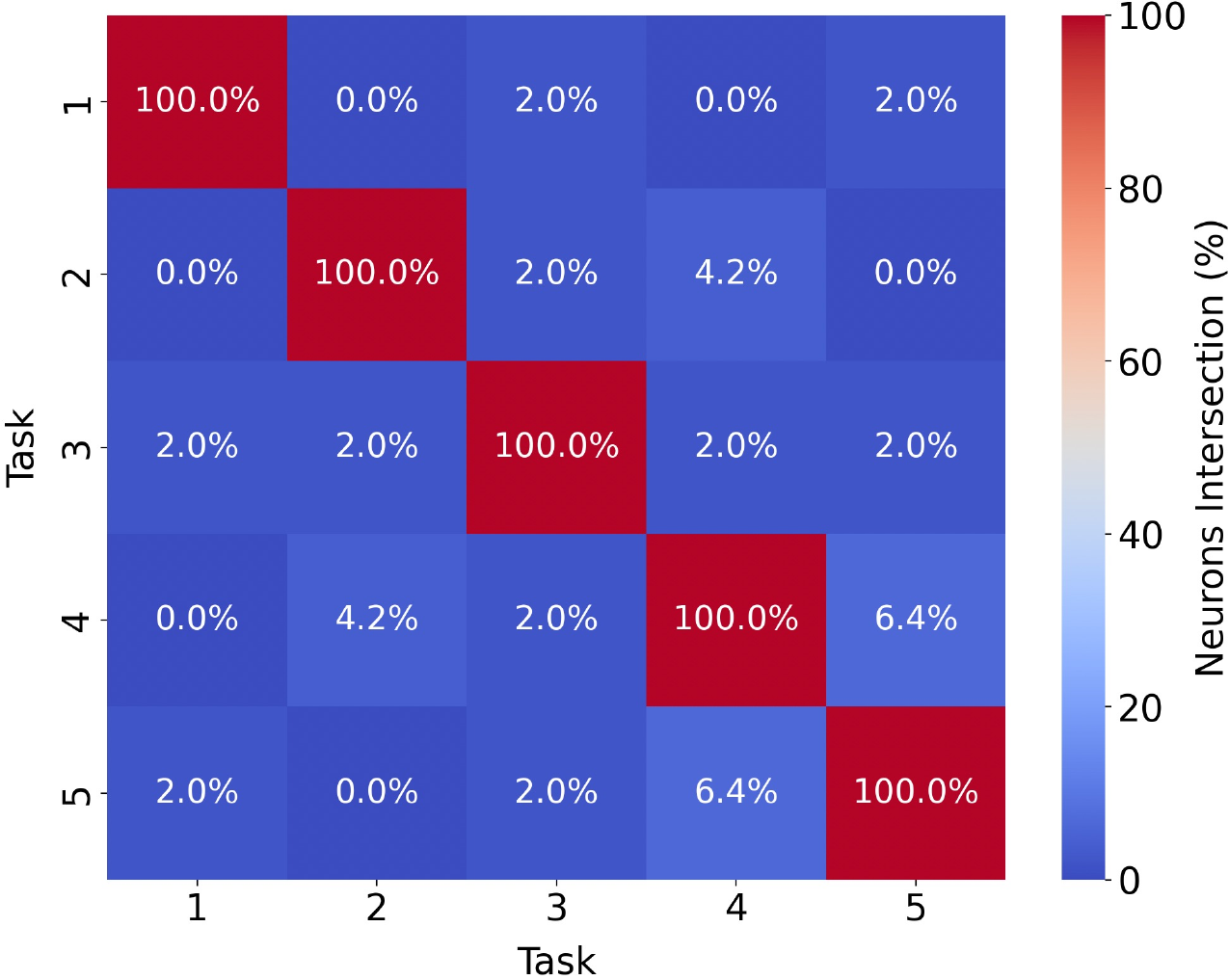}
  \vspace{-0.1in}
  \caption{Neuron overlap heatmap showing the percentage of neuron intersection across five tasks. Each cell represents the normalized intersection-over-union score between two tasks. Diagonal values of 100\% in self-comparison highlight that distinct neurons are exclusively utilized for different tasks.}
\label{fig:neurons-relied-tasks}
  \vspace{-0.0in}
\end{figure}

To further investigate the root cause of the observed neuron stability, we quantified the overlap among neurons utilized during the training of individual tasks using the normalized intersection-over-union (IoU) score and visualized these intersections as a heatmap in Figure~\ref{fig:neurons-relied-tasks}. The analysis revealed that the largest intersection, excluding intra-task comparisons, is 6.4\%, observed between Tasks 4 and 5, suggesting a small degree of overlap. However, the overall low IoU scores indicate that tasks primarily rely on distinct sets of neurons.
These results demonstrate that different tasks in Cl rely on predominantly disjoint sets of neurons, and the neurons critical to a task remain stable in learning subsequent tasks.

%

\vspace{-0.1in}
\section{Persistent Backdoor Attack}
\label{sec: methodology}
In this section, we introduce the threat model, including two adversaries with different capabilities and key properties, followed by an overview of the proposed persistent backdoor.

\begin{figure*}[t]
 \centering
  \includegraphics[width=0.95\textwidth]{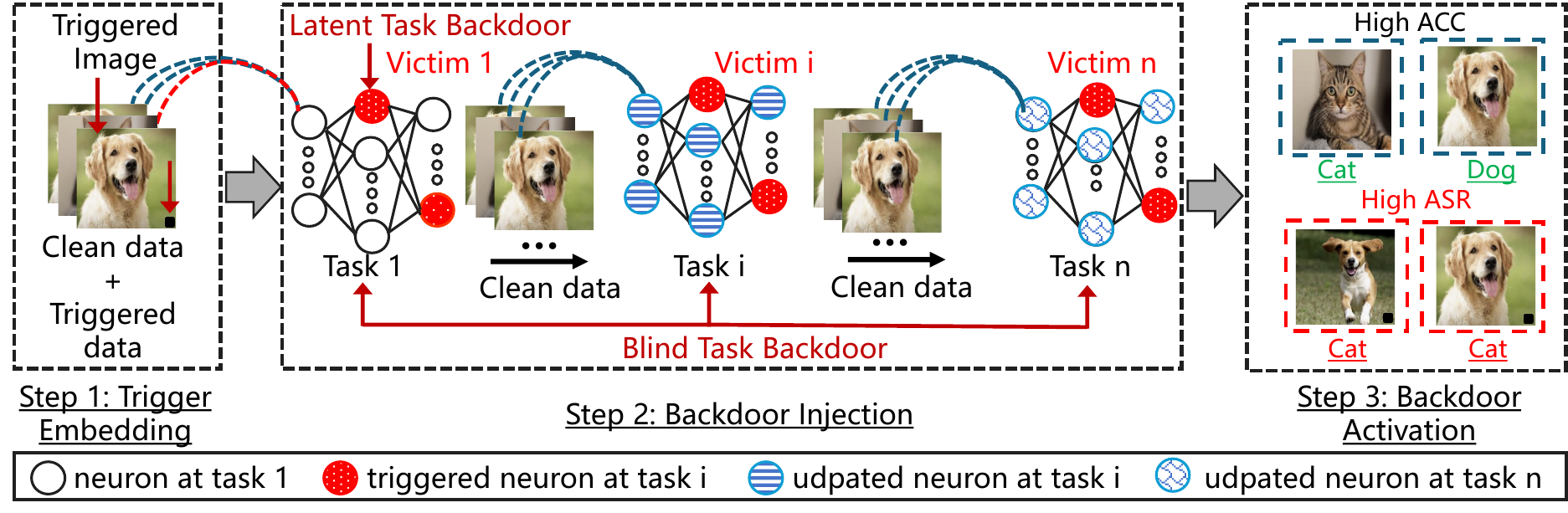} 
  \vspace{-0.1in}
  \caption{The three steps of our Persistent Backdoor Attack. \textit{Step 1: Trigger Embedding}--This involves generating triggered training data during the training process of a specific task. \textit{Step 2: Backdoor Injection}--The attack code, generated by either \textbf{blind backdoor} or \textbf{latent backdoor}, implants the backdoor into the model during training. \textit{Step 3: Backdoor Activation}--When fed with the triggered input, the backdoored model produces an incorrect prediction, which persists across all subsequent tasks.}
  \label{fig:attack-three-steps}
  \vspace{-0.15in}
\end{figure*}

\vspace{-0.1in}
\subsection{Threat Model and Assumptions}
\label{sec: threatmodel}
%
%
%
The rise in deep learning's popularity has turned platforms like GitHub and Hugging Face into repositories for hundreds of thousands of open-source models. These models are publicly available either as pre-trained weights, ready for immediate use in various tasks, or as source code, which developers can adapt for specific applications. This accessibility has led to the widespread integration of open-source code, driven by the need to reduce costs and accelerate the development of large models~\cite{synopsys2023ossra}. For instance, Hugging Face reported over 600 million model downloads in August 2023 alone~\cite{stateofai2023}. 
However, due to the widespread adoption of platforms like GitHub, Twitter, and Flicker in the ML community, the attack vector has expanded beyond poisoned data to include poisoned weights and carefully crafted code snippets~\cite{SahSubPir20,bagd-2021}.
Given such threats, we consider two specific attacks--\textit{\textbf{\spba}} and \textit{\textbf{\wpba}}--in which the adversary targets backdooring a specific task. These threats are stealthy, as the backdoor remains dormant until the targeted task is executed, significantly reducing the risk of their detection.

The blind task backdoor (BTB), which we will refer to as the \textit{blind backdoor} for brevity, is a type of code poisoning attack where the adversary indirectly impacts the model training process without access to the training data. Similar to~\cite{bagd-2021}, the adversary maliciously manipulates the machine learning codebase, referred to as the \textit{attack code}, specifically targeting the calculation of the loss value during training. The attack code also includes a module for generating poisoned data samples by injecting triggers into a small subset of the training data and assigning incorrect labels, as detailed in~\cite{bagd-2021}. As a result, the adversary remains blind to both the training process and the data. In continual learning, the same attack code is used across all tasks, allowing the backdoor to be injected at any point in the learning process without the adversary directly interacting with the model during training. This attack is particularly effective when ML code sourced from a repository is used unchanged to train a model, allowing the adversary to embed malicious behaviors without the developer's knowledge, thereby compromising the application's integrity.
We note that loss computation in deep learning models is context-dependent, varying with architecture, data modality, tasks, and the CL algorithm. This diversity often results in a single codebase employing multiple loss functions, making it challenging to detect malicious alterations through code review alone, as detailed in~\cite{bagd-2021}.

Unlike the blind backdoor, which relies on indirect manipulation of the entire training process, the latent task backdoor (LTB), referred to as the \textit{latent backdoor}, is a more sophisticated attack method in which the adversary manipulates the training process of \textit{only one task}. The weight modifications introduced during this task are designed to backdoor future targeted tasks, even when their training processes remain benign.  
This restriction on the adversary showcases that, in CL, as in classical ML pipelines, \textit{publicly shared weights can contain dormant and hidden functionalities that may be exploited in later tasks}. The adversary achieves this by modifying the loss computation for a single task, leading to a controlled update of specific weights during the training process.
The restricted threat model of LTBs (\ie single-task manipulation) highlights the risks developers face when directly using pre-trained weights to build applications, fine-tune custom models, few-shot learning~\cite{SunLiuChu19}, or perform transfer learning~\cite{YaoLiZhe19}. It also underscores that backdooring poses a significant threat in pipelines relying on untrusted foundation model weights for learning additional tasks~\cite{YanZhoDin24}, \textit{even when the training pipeline itself is verified and trusted}. 

We emphasize that in scenarios where the training pipeline is developed by legitimate entities and remains benign, the blind backdoor becomes inapplicable, as it relies on manipulating the loss computation through malicious code. Specifically, in a benign training pipeline, where the loss computation is not compromised, the blind backdoor cannot function as intended, even if the sourced pre-trained weights are backdoored. In contrast, our proposed latent backdoor remains effective by leveraging controlled manipulation of a single task during pre-training, embedding a dormant backdoor that activates during the benign training of subsequent tasks.
Finally, we note that detecting changes in pre-trained weights is challenging due to the randomness introduced by factors like the optimizer, initial seed, and data variations. In Section~\ref{sec: evaluation}, we present empirical results on how these attacks evade backdoor detection mechanisms.

In summary, for BTB, we assume that the adversary compromises the loss computation in the ML code. Although the adversary may have knowledge of the architecture or data domain, they lack access to the training data and have no knowledge of the hyperparameters. In contrast, for LTB, the adversary controls the training process of a single task.
\subsection{Attack Overview}
\label{subsec:attack-overview}
%
%
Conducting the proposed persistent backdoor attack, both blind and latent, involves three major steps (Figure~\ref{fig:attack-three-steps}): \textit{(i)} embedding a trigger into a small subset of training data samples; \textit{(ii)} injecting the backdoor into the attack model during the training process to build a backdoored model;  
\textit{(iii)} activating the backdoor during inference, in which malicious actors feed triggered images to the backdoored model, causing incorrect predictions. Here, we will elaborate on these steps.

\noindent
\underline{\textbf{Step 1: Trigger Embedding.}} In backdoor attacks, trigger embedding involves selecting a small subset of the training data (4\% in our experiments) and inserting a hidden pattern into each sample within this subset. The labels of these samples are then switched to an incorrect target label. Various trigger types exist, including static (fixed shape and location), dynamic (variable shape and location), physical (an object in the environment), and text-based triggers (a specific word(s)). We demonstrate the power of persistent attacks using these triggers. In our proposed persistent backdoor, trigger embedding is performed on the fly through a module of the attack code, without the adversary's access to the training data.

\noindent
\underline{\textbf{Step 2: Backdoor Injection.}} A successful backdoor attack in continual learning must satisfy three primary requirements: \textit{(i)} achieving a high attack success rate on the triggered data samples in the current task, \textit{(ii)} maintaining a high attack success rate in tasks following the backdoor injection, and \textit{(iii)} preserving high target classification accuracy on clean data across all tasks, both before and after backdoor activation. However, simultaneously satisfying these conditions is challenging. As the model's weights evolve to incorporate new knowledge, some information, including the embedded trigger, may be erased (as in Figure~\ref{motivation-fig-1}), diminishing the attack's success rate.
To address this challenge, we propose two persistent backdoor attacks, customized for continual learning. The first is the \spba attack, where the adversary manipulates the loss function in the ML code, impacting the loss value computation of all tasks. A \textbf{key innovation} in this design is conceptualizing the backdoor attack as a multi-objective optimization problem. This approach requires balancing several objectives, \ie maximizing the attack success rate with triggered data, and maintaining high classification accuracy before and after tasks, to achieve Pareto optimality. In doing so, our design strategically manages trade-offs between these objectives to find the most effective balance~\cite{Jedlicka2022ParetoOE}.
Our blind backdoor attack effectively addresses this challenge, as detailed in Section~\ref{subsection:saa}. 

While the blind backdoor enforces an altered loss function for all tasks, it raises an important question: \textit{Can an attack succeed by compromising the loss value computation of just a single task?} This consideration leads to the design of the \wpba; a targeted backdoor attack where the adversary manipulates the loss calculation for a single target task, leaving all prior and future tasks seemingly benign. The details of this attack are discussed in Section~\ref{subsection:weakerattack}.
%

\noindent
\underline{\textbf{Step 3: Backdoor Activation.}} During this phase, the backdoored model is deployed for inference. Under normal circumstances, when fed benign input data, the model behaves as expected and predicts correct labels. However, to execute the attack, the adversary introduces a triggered input--a sample containing the trigger pattern similar to the one used in training--which causes the model to predict an incorrect target label pre-set by the attacker (as illustrated in Step 3 of Figure~\ref{fig:attack-three-steps}). In the context of continual learning, we show our proposed blind and latent backdoor attacks demonstrate persistence, with the malicious behavior carrying over into all subsequent tasks learned by the model.
\section{Attack Design}
\label{sec:attack-details}
%

\subsection{\saa}
\label{subsection:saa}
As mentioned in Section~\ref{sec: threatmodel}, in \spba (\ie blind backdoor), the adversary generates the attack code by manipulating the loss value computation of the ML code, indirectly impacting the training process of the model across all tasks. For this attack to be successful, the modification of the loss computation should achieve a high attack success rate on the triggered inputs on the current and future tasks, and maintain high classification accuracy on the clean inputs across all tasks. 
To achieve this, we devise a customized blind loss, aiming to effectively solve this multi-objective optimization problem, as detailed in Eqn.~\ref{eqn:loss}: 
%
\vspace{-0.04in}
\begin{equation}
\label{eqn:loss}
\ell_{blind} = L(\theta(x), y) + \lambda L(\theta(x^+), y^+),
\end{equation}
in which $L$ represents the loss criterion, $\theta$ denotes the model parameters, $x$ and $y$ are the features and ground truth labels of clean data samples, $x^+$ and $y^+$ are the features and ground truth labels of the triggered data samples, and $\lambda$ is the penalty coefficient.
The blind loss $\ell_{blind}$ is a linear combination of the task loss, $L(\theta(x), y)$, which computes the loss value of the clean data, and the backdoor loss, $L(\theta(x^+), y^+)$, which computes the loss value of the triggered data.

To maintain the stealth of backdoor attacks, the triggered dataset is typically much smaller than the clean dataset to limit the extent of training data modifications, leading to data imbalance. However, such data imbalances undermine the backdoor attack by shifting the model's focus toward learning the clean data, diminishing the effectiveness of the triggered samples. 
To address this challenge, we introduce a penalty coefficient $\lambda$ to the backdoor loss. This coefficient imposes an additional penalty whenever the model incorrectly predicts the target label (the label chosen by the attacker) for a triggered sample, ensuring effective attack and classification. 
%
\setcounter{algorithm}{0}
\begin{algorithm}[t]
\caption{\saa (BTB)}
\label{alg: stronger-attack-algorithm}
\begin{algorithmic}[1] 

\Statex \textbf{Input:} $\theta$ (weights),  $n$ (iteration number), $\alpha$ (learning rate ), $\beta$ (regularization), $\tau$ (tolerance threshold)  

\setcounter{ALG@line}{0} 
\State  $\lambda \gets 0$, $\mu \gets 0.1$, $\gamma \gets 0.99$
\For{$k = 1$ to $n$} 
    \State $\theta \leftarrow \theta - \alpha \nabla_{\theta} L$ \hspace{0.1in} \textcolor{SeaGreen}{\(\triangleright\) $L$ is the loss function}
    \State $\lambda \leftarrow \lambda + \beta \left[ \ell_{blind}^j - \tilde{\ell}_{blind}^j - \tau \right]$ for each previous $j$ 
    \State $\mu \leftarrow \mu \cdot \gamma$ 

\EndFor
\State \textbf{return} $\theta$ 
\end{algorithmic}
\end{algorithm}

In continual learning, updating the model's parameters for the current task, $i$ can impact the loss value of a previous task $j$, resulting in an updated loss $\widetilde{\ell}_{blind}^{j}$, which may degrade the model's classification accuracy on prior learned tasks. This degradation poses a significant challenge to our attack objectives, which include maintaining high classification accuracy across all tasks and ensuring persistent attack efficacy. To address this challenge, the blind backdoor controls the loss calculation $\ell_{blind}^i$ for the current task by constraining the variation of the preceding task's loss $\ell_{blind}^j$ within a specified threshold, $\tau$:
\vspace{-0.03in}
\begin{equation}
\begin{aligned}
   & \underset{\theta_i}{\min} \, (L(\theta_i(x), y) + \lambda L(\theta_i(x^+), y^+) ) \\ 
   & \text{s.t.} \hspace{5mm} \widetilde{\ell}_{blind}^{j} \leq \ell_{blind}^{j} + \tau, \hspace{0.1in} \forall j \in \{1,...,i-1\}. 
\end{aligned}
\label{eq:loss_func}
\end{equation}

Finally, to solve this multi-objective optimization problem, we employ the \textbf{augmented Lagrangian} approach. This method incorporates the constraints into the loss function by introducing Lagrange multipliers and penalty terms, seamlessly integrating constraint satisfaction into the optimization process. 
To optimize the model parameters $\theta$, we minimize the value of Eqn.~\ref{eq:augmented_lagrangian} using gradient descent, iterating up to a maximum of $n$ iterations. Algorithm~\ref{alg: stronger-attack-algorithm} depicts the process of the blind task backdoor. Lines 3-5 of the algorithm update the model weights $\theta$, the Lagrange multipliers $\lambda$, and the penalty coefficient $\mu$. If the early stopping condition specified in line 5 is met, the algorithm terminates prematurely. 

\begin{algorithm}[t]
\caption{\waa (LTB)}
\label{alg: weaker_algorithm}
\begin{algorithmic}[1] 
\Statex \textbf{Input:} $D^{clean}$ (clean dataset), $D^{trigger}$ (triggered dataset), $v^{trigger}$ (trigger embedding value), $\kappa$ (importance threshold), $P$ (selection percentage), $\epsilon$ (tolerance threshold)

\setcounter{ALG@line}{0} 

\Statex \hspace{-0.25in} \textcolor{SeaGreen}{\(\triangleright\) Compute Diagonal Fisher Matrix}
\State $F \gets []$
\For{$x, y$ in $D^{clean}$}
    \State $\ell \gets L(\theta(x), y)$
    \State Compute $F \gets \nabla_{\theta}^2 \ell$ for each $\theta$
\EndFor

\Statex \hspace{-0.25in} \textcolor{SeaGreen}{\(\triangleright\) Select Stable Neurons from \(n\) Neurons}
\State $imps = \{z \mid z \in \theta \text{ and } F[z] \geq \kappa \},\ \text{s.t.} \ |imps| = P \times n$

\Statex \hspace{-0.25in} \textcolor{SeaGreen}{\(\triangleright\) Train with Triggered Data} 
\For{$x^+, y^+$ in $D^{trigger}$}
    \State Update $\theta$ with embedded $imps$ using $v^{trigger}$
    \State $\ell_{latent} \gets L_i(\theta(x^+), y^+) + \text{ReLU}(L_i(\theta(x), y) - \epsilon)$
    \State Backpropagate: $\ell_{latent}.backward()$
\EndFor
\State \textbf{return} $\theta$
\end{algorithmic}
\end{algorithm}

\vspace{-0.04in}
\begin{equation}
\begin{aligned}
& \underset{\theta, \lambda, \mu}{\min} \, (\ell_{blind}^{i} + \sum_j \lambda_j \delta  + \frac{\mu}{2} \sum_j \delta^2), 
\\
& \hspace{2mm}  \text{s.t.} \hspace{2mm} \delta = \ell_{blind}^j - \widetilde{\ell}_{blind}^{j} - \tau.
\end{aligned}
\label{eq:augmented_lagrangian}
\end{equation}
\vspace{-0.2in}

\begin{table*}[t]
  \centering
  \caption{Classification accuracy of various continual learning algorithms across different datasets and architectures.}
  \vspace{-0.1in}
  \footnotesize
  \begin{tabular}{c c c c c c c c}
    \toprule
    Model & Dataset & SI~\cite{si2017} & EWC~\cite{ewc2016} & XdG~\cite{xdg2019} & LwF~\cite{lwf2017} & DGR~\cite{dgr2018} & A-GEM~\cite{agem2019} \\
    \midrule
    \multirow{4}{*}{CNN} & SplitMNIST & 0.99 $\pm$ 0.01 & 1.0 $\pm$ 0.01 & 0.94 $\pm$ 0.02 & 1.0 $\pm$ 0.01 & 1.0 $\pm$ 0.01 & 1.0 $\pm$ 0.01 \\
    \cmidrule(lr){2-8}
          & PermutedMNIST & 0.85 $\pm$ 0.02 & 0.90 $\pm$ 0.01 & 0.83 $\pm$ 0.03 & 0.89 $\pm$ 0.02 & \textbf{0.22} $\pm$ 0.06 & 0.88 $\pm$ 0.02 \\
    \cmidrule(lr){2-8}
          & SplitCIFAR10 & 0.79 $\pm$ 0.02 & 0.85 $\pm$ 0.01 & 0.86 $\pm$ 0.02 & 0.85 $\pm$ 0.03 & \textbf{0.59} $\pm$ 0.05 & 0.88 $\pm$ 0.01 \\
    \midrule
    \multirow{4}{*}{ResNet18} & SplitMNIST & 0.99 $\pm$ 0.01 & 1.0 $\pm$ 0.01 & 0.94 $\pm$ 0.02 & 1.0 $\pm$ 0.01 & 1.0 $\pm$ 0.01 & 1.0 $\pm$ 0.01 \\
    \cmidrule(lr){2-8}
          & PermutedMNIST & 0.87 $\pm$ 0.02 & 0.92 $\pm$ 0.02 & 0.84 $\pm$ 0.02 & 0.93 $\pm$ 0.02 & \textbf{0.37} $\pm$ 0.05 & 0.89 $\pm$ 0.01 \\
    \cmidrule(lr){2-8}
          & SplitCIFAR10 & 0.81 $\pm$ 0.02 & 0.86 $\pm$ 0.01 & 0.86 $\pm$ 0.02 & 0.88 $\pm$ 0.01 & \textbf{0.64} $\pm$ 0.04 & 0.91 $\pm$ 0.01 \\
    \bottomrule
  \end{tabular}
  \label{tab:without-attack-acc}
\vspace{-0.05in}
\end{table*}

\subsection{\waa}
\label{subsection:weakerattack}
In the latent backdoor attack, the adversary, in control of the training process for a single task (\eg task $i$), modifies the loss computation exclusively for that specific task. The \textbf{key idea} in constructing the latent backdoor attack is to embed the backdoor into the model's \textit{most stable components,} ensuring that the \textit{malicious behavior propagates into all subsequent tasks,} despite their benign training (both ML code and process). As a result, this malicious behavior remains latent until the trigger is activated, causing any triggered input to be classified as the target label, chosen by the attacker.

The primary challenge in latent task backdooring is ensuring that the trigger remains effective and activates as intended in future tasks without being diminished. 
To tackle this challenge, we design a customized latent loss as a linear combination of the backdoor loss and the task loss:
\vspace{-0.04in}
\begin{equation}
\begin{aligned}
\ell_{latent}^i = L_i(\theta(x^+), y^+) + Relu(L_i(\theta(x), y)- \epsilon).
\end{aligned}
\label{eqn:ltb_loss_func}
\vspace{-0.04in}
\end{equation}
In Eqn.~\ref{eqn:ltb_loss_func}, $L_i$ represents the loss criterion of task $i$, $\theta$ denotes the model parameters, $x^+$ and $y^+$ refer to the features and labels of the triggered samples, and $x$ and $y$ refer to the features and labels of the clean data. The threshold $\epsilon$ is a hyperparameter used to maintain the classification accuracy of the clean data and is set as a factor of the backdoor loss (empirically set to 0.1 of $L_i(\theta(x^+), y^+)$ in our experiments). The term $Relu(L_i(\theta(x), y) - \epsilon)$ ensures that the classification performance on the clean data stays within an acceptable range. Specifically, if $L_i(\theta(x), y)$ is less than $\epsilon$, the loss value for the clear data is set to zero. otherwise, the loss will be $L_i(\theta(x), y) - \epsilon$.

The next step in the process involves selecting neurons for trigger embedding. Our neuron-level analysis in Section~\ref{sec: Observation} suggests that embedding the trigger in the most stable neurons is ideal for ensuring attack persistence. However, as outlined in our threat model, the LTB adversary's access to only one task poses a significant challenge to analyzing neuron stability across multiple tasks.
We addressed this challenge by shifting our focus to assessing neuron significance. Specifically, as depicted in Algorithm~\ref{alg: weaker_algorithm}, we employ the Diagonal Fisher Matrix (DFM) to evaluate neuron significance and derive the importance matrix $F$ (lines 1–4). We chose DFM for its computational efficiency compared to the Fisher Importance Matrix (FIM). Unlike FIM, which imposes high computational complexity due to the covariance computation across all neuron pairs, DFM focuses solely on the diagonal elements--each neuron's individual contribution to the model's output--significantly reducing computational demands. 

DFM analysis enables us to evaluate neurons' contributions to the classification performance of the task controlled by the adversary by assessing the loss function's sensitivity to variations in neurons.
Interestingly, our in-depth analysis revealed a significant overlap between the stable neurons and those identified by DFM analysis as crucial for the task's classification performance. As a result, we select the top important neurons based on the importance matrix $F$, forming the candidate set, with $\kappa$ approximately corresponding to the 98th percentile of $F$ in our experiments.
%
To maintain the attack's stealth and persistence, we select a proportion $P\%$ of the candidate set for trigger embedding (line~5). Our algorithmic-level observations in Section~\ref{sec: Observation} revealed that replay-based algorithms experience more significant parameter changes, necessitating a higher percentage of neurons for trigger embedding in replay-based algorithms. In our experiments, we set $P$ to 70\% and 90\% of the candidate set for regularization-based and replay-based algorithms, respectively.

The final step is to train the model using the triggered data. To ensure the persistence of the embedded latent backdoor in subsequent tasks, we embed the backdoor behavior by adding a \textbf{small and controlled} value, $v^{trigger}$, to the weights of the selected neurons. This operation is carefully designed to create the malicious behavior that activates only when the input data contains the trigger pattern. During training, the triggered information is embedded into the selected neurons by incrementally adjusting their values using $v^{trigger}$. Upon completing the training for task $i$ on the triggered dataset, we obtain the updated model parameters $\theta$ (lines~6-10). 


\section{Experiments}
\label{sec: evaluation}
To validate the effectiveness of our proposed blind backdoor (\textit{\textbf{BTB}}) and latent backdoor (\textit{\textbf{LTB}}) attacks, we conducted extensive experiments (details are in Table~\ref{tab:experiment-configuration} in the Appendix) to evaluate: (\textit{i}) static backdoor against six CL algorithms, (\textit{ii}) dynamic backdoor (\textit{iii}) physical backdoor, (\textit{iv}) backdoor on text classification, (\textit{v}) comparative analysis against other backdoor methods, (\textit{vi}) backdooring different tasks in CL, and (\textit{vii}) evading existing backdoor detection defenses\footnote{Code is available on \url{https://doi.org/10.5281/zenodo.14728872}}.
We perform backdoor attacks on task-based CL with the code developed based on~\cite{vandeven2022three}. 
The backdoor is embedded in task~1 unless stated otherwise. All experiments are conducted using PyTorch and executed on an NVIDIA A100 GPU from ACCESS CI~\cite{Access2023}. 
For BTB, we set the number of iterations ($n$) to 300, $\alpha$ to $0.001$, $\beta$ to $0.0001$, and $\tau$ to be $0.05 \times \ell_j$, where $\ell_j$ is the loss value of original task $j$. For the LTB attack, we set $v^{tigger}$ to 0.5, $\epsilon$ to 0.1 of the backdoor loss ($0.1 \times L(\theta(x^+),y^+)$), $\kappa$ to 98 percentile of all neurons, and $P$ to 70\% and 90\% for regularization-based and replay-based algorithms.

\begin{table*}[t]
  \centering
  \caption{{\it \wpba} attack evaluation using static trigger across all algorithms, datasets, and architectures.}
\vspace{-0.1in}
  \resizebox{\textwidth}{!}{
  \begin{tabular}{
>{\centering\arraybackslash}m{3em} 
>{\centering\arraybackslash}m{6em} 
>{\centering\arraybackslash}m{3em} 
>{\centering\arraybackslash}m{3em} 
>{\centering\arraybackslash}m{3em} 
>{\centering\arraybackslash}m{3em} 
>{\centering\arraybackslash}m{3em} 
>{\centering\arraybackslash}m{3em} 
>{\centering\arraybackslash}m{3em} 
>{\centering\arraybackslash}m{3em} 
>{\centering\arraybackslash}m{3em} 
>{\centering\arraybackslash}m{3em} 
>{\centering\arraybackslash}m{3em} 
>{\centering\arraybackslash}m{3em}}
    \toprule
    \multirow{2}{*}{Model} & \multirow{2}{*}{Dataset} & \multicolumn{2}{c}{SI} & \multicolumn{2}{c}{EWC} & \multicolumn{2}{c}{XdG} & \multicolumn{2}{c}{LwF} &\multicolumn{2}{c}{DGR} & \multicolumn{2}{c}{A-GEM} \\
    \cmidrule(lr){3-4} \cmidrule(lr){5-6} \cmidrule(lr){7-8} \cmidrule(lr){9-10} \cmidrule(lr){11-12} \cmidrule(lr){13-14}
    & & ASR & ACC & ASR & ACC & ASR & ACC & ASR & ACC & ASR & ACC & ASR & ACC \\
    \midrule
    \multirow{6}{*}{CNN} & \multirow{1.5}{*}{SplitMNIST} & 
    \makecell{$0.95$ \\ $\pm 0.05$} & 
    \makecell{$0.92$ \\ $\pm 0.02$} & 
    \makecell{$0.93$ \\ $\pm 0.03$} & 
    \makecell{$0.91$ \\ $\pm 0.02$} & 
    \makecell{$0.94$ \\ $\pm 0.01$} & 
    \makecell{$0.91$ \\ $\pm 0.03$} & 
    \makecell{$0.91$ \\ $\pm 0.05$} & 
    \makecell{$0.92$ \\ $\pm 0.02$} &
    \makecell{$0.92$ \\ $\pm 0.04$} & 
    \makecell{$0.93$ \\ $\pm 0.03$} & 
    \makecell{$0.91$ \\ $\pm 0.07$} & 
    \makecell{$0.93$ \\ $\pm 0.03$} \\
    \cmidrule(lr){2-14}          
    & \multirow{1.5}{*}{PermutedMNIST} &  
    \makecell{$0.93$ \\ $\pm 0.04$} & 
    \makecell{$0.82$ \\ $\pm 0.09$} & 
    \makecell{$0.85$ \\ $\pm 0.09$} & 
    \makecell{$0.80$ \\ $\pm 0.05$} & 
    \makecell{$0.90$ \\ $\pm 0.03$} & 
    \makecell{$0.82$ \\ $\pm 0.01$} & 
    \makecell{$0.92$ \\ $\pm 0.06$} & 
    \makecell{$0.85$ \\ $\pm 0.03$} &
    \makecell{$0.91$ \\ $\pm 0.04$} & 
    \makecell{0.26 \\ $\pm 0.02$} & 
    \makecell{$0.89$ \\ $\pm 0.03$} & 
    \makecell{$0.81$ \\ $\pm 0.04$} \\
    \cmidrule(lr){2-14}          
    & \multirow{1.5}{*}{SplitCIFAR10} & 
    \makecell{$0.91$ \\ $\pm 0.02$} & 
    \makecell{$0.76$ \\ $\pm 0.01$} & 
    \makecell{$0.91$ \\ $\pm 0.03$} & 
    \makecell{$0.83$ \\ $\pm 0.04$} &  
    \makecell{$0.88$ \\ $\pm 0.04$} & 
    \makecell{$0.85$ \\ $\pm 0.01$} & 
    \makecell{$0.84$ \\ $\pm 0.04$} & 
    \makecell{$0.81$ \\ $\pm 0.03$} &
    \makecell{$0.94$ \\ $\pm 0.03$} & 
    \makecell{0.52 \\ $\pm 0.02$} & 
    \makecell{$0.89$ \\ $\pm 0.06$} & 
    \makecell{$0.86$ \\ $\pm 0.03$} \\
    \midrule
    \multirow{6}{*}{ResNet18} & \multirow{1.5}{*}{SplitMNIST} & 
    \makecell{$0.93$ \\ $\pm 0.01$} & 
    \makecell{$0.96$ \\ $\pm 0.03$} & 
    \makecell{$0.91$ \\ $\pm 0.01$} & 
    \makecell{$0.92$ \\ $\pm 0.04$} & 
    \makecell{$0.93$ \\ $\pm 0.03$} & 
    \makecell{$0.91$ \\ $\pm 0.02$} & 
    \makecell{$0.91$ \\ $\pm 0.01$} & 
    \makecell{$0.98$ \\ $\pm 0.04$} &
    \makecell{$0.89$ \\ $\pm 0.03$} & 
    \makecell{$0.91$ \\ $\pm 0.02$} & 
    \makecell{$0.90$ \\ $\pm 0.03$} & 
    \makecell{$0.93$ \\ $\pm 0.04$}  \\
    \cmidrule(lr){2-14}          
    & \multirow{1.5}{*}{PermutedMNIST} &  
    \makecell{$0.93$ \\ $\pm 0.01$} & 
    \makecell{$0.86$ \\ $\pm 0.06$} & 
    \makecell{$0.89$ \\ $\pm 0.03$} & 
    \makecell{$0.88$ \\ $\pm 0.04$} & 
    \makecell{$0.92$ \\ $\pm 0.01$} & 
    \makecell{$0.79$ \\ $\pm 0.03$} & 
    \makecell{$0.93$ \\ $\pm 0.06$} & 
    \makecell{$0.91$ \\ $\pm 0.07$} &
    \makecell{$0.91$ \\ $\pm 0.01$} & 
    \makecell{0.35 \\ $\pm 0.06$} & 
    \makecell{$0.85$ \\ $\pm 0.03$} & 
    \makecell{$0.85$ \\ $\pm 0.01$} \\
    \cmidrule(lr){2-14}          
    & \multirow{1.5}{*}{SplitCIFAR10} & 
    \makecell{$0.90$ \\ $\pm 0.03$} & 
    \makecell{$0.80$ \\ $\pm 0.04$} & 
    \makecell{$0.90$ \\ $\pm 0.03$} & 
    \makecell{$0.82$ \\ $\pm 0.02$} &  
    \makecell{$0.82$ \\ $\pm 0.02$} & 
    \makecell{$0.78$ \\ $\pm 0.03$} & 
    \makecell{$0.90$ \\ $\pm 0.01$} & 
    \makecell{$0.83$ \\ $\pm 0.03$} &
    \makecell{$0.89$ \\ $\pm 0.02$} & 
    \makecell{0.63 \\ $\pm 0.03$} & 
    \makecell{$0.93$ \\ $\pm 0.01$} & 
    \makecell{$0.88$ \\ $\pm 0.07$} \\
    \bottomrule
  \end{tabular}}
  \label{tab:static-waa}
  \vspace{-0.15in}
\end{table*}

\subsection{Datasets and Classification Models}
\label{subsec:dataset}
\textit{\textbf{Dataset.}} We use (\textit{i}) SplitMNIST for five tasks, each containing two categories, with 12,000 samples for training and 2,000 samples for testing~\cite{vandeven2022three}; (\textit{ii}) PermutedMNIST for ten tasks, each including all ten digit categories, with 60,000 samples for training and 10,000 samples for testing~\cite{si2017}; and (\textit{iii}) SplitCIFAR10 for five tasks, each containing two categories, with 10,000 samples for training and 2,000 samples for testing~\cite{vandeven2022three}.

\smallskip
\noindent
\textit{\textbf{Architecture.}} We use two architectures:(\textit{i}) a five-layer CNN architecture with five standard convolution layers with two $3\times3$ filters and ReLU activation function, followed by three fully connected layers; and (\textit{ii}) a ResNet18. Refer to Table~\ref{tab:three-neural-architectures} in the Appendix for the details of these architectures.
%

\smallskip
\noindent
\textit{\textbf{CL Algorithm Performance Analysis.}} We first conduct experiments to assess the performance of six CL algorithms using three clean datasets. Table~\ref{tab:without-attack-acc} presents the average classification accuracy (ACC) across different models and datasets. 
Overall, the results indicate that most algorithms maintain high ACC across all datasets and architectures. EWC outperforms other algorithms on PermutedMNIST, while A-GEM achieves the best performance on SplitCIFAR10. Given the lower complexity of the SplitMNIST dataset, all algorithms achieve above 94\% ACC, demonstrating their effectiveness in simpler scenarios.
%
%
However, we observed that DGR is the least-performing algorithm across different models and datasets, except for SplitMNIST. Notably, its performance drops significantly to an average of 45.5\% on PermutedMNIST and SplitCIFAR10. This decline is attributed to DGR's heavy reliance on the quality of its generator~\cite{xdg2019}. As a result, DGR struggles with complex datasets or a larger number of tasks, making it particularly vulnerable to catastrophic forgetting.

\subsection{Static Backdoor Attack}
\label{subsec:static}
Next, we evaluate the persistence of our proposed LTB and BTB attacks using static triggers. 

\smallskip
\noindent
\textit{\textbf{Main task.}} We conducted this experiment on all six CL algorithms, using the CNN and ResNet18 architectures across all datasets; resulting in a total of thirty-six attack combinations for each attack type. We trained the model for 100 epochs using the Adam optimizer with a batch size of 128 and a learning rate of 0.001. 

\smallskip
\noindent
\textbf{Static-backdoor Task.} For this experiment, we implanted the backdoor in the first task by selecting 5\% of the training samples. We inserted a $4\times4$-pixel static trigger, with pixel values set to 0, at the bottom right corner. 

\begin{figure}[t]
\vspace{-0.05in}
  \centering
  \includegraphics[width=\columnwidth]{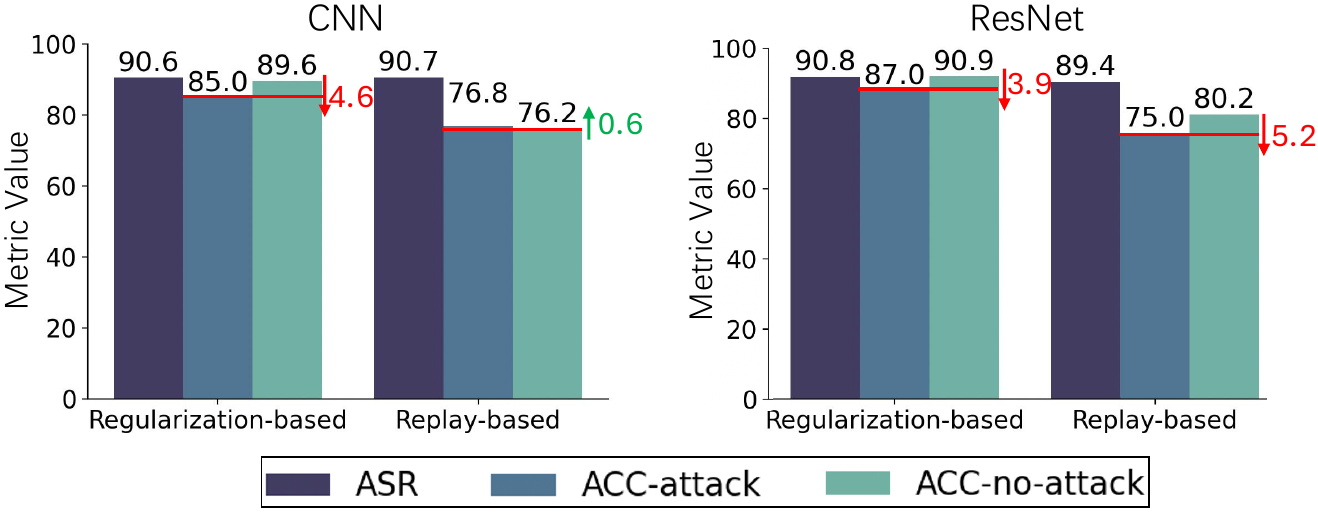}
  \vspace{-0.25in}
  \caption{Performance analysis of LTB attack using the static trigger and its impact on classification accuracy for CNN (left) and ResNet18 (right) architectures. We calculated two separate averages: one across all regularization-based algorithms and another across all replay-based algorithms. The LTB attack achieves high ASR across all combinations with only a negligible drop in accuracy.} 
  \label{fig:static_attack_vs_without_attack}
  \vspace{-0.0in}
\end{figure}

\smallskip
\noindent
\textit{\textbf{Results.}} Table~\ref{tab:static-waa} presents the results of the LTB attack (refer to Table~\ref{tab:static-saa} in the Appendix for BTB attack results), indicating the effectiveness of our latent backdoor against all CL algorithms; the blind backdoor exhibits similar attack efficacy. 
In most scenarios, the proposed LTB attack achieves attack success rates (ASRs) above \textbf{90\%}, with the lowest ASR being 82.02\% for the attack on ResNet18-SplitCIFAR10 using XdG.
We summarized these results in Figure~\ref{fig:static_attack_vs_without_attack}. One can observe that when using CNN architecture, the LTB attack against regularization-based and replay-based algorithms achieves an average ASR of \textbf{90.6\%} and \textbf{90.7\%}, respectively. For the CNN architecture, the LTB attack has led to minimal accuracy drops of \textbf{4.6\%} for regularization-based algorithms, but a negligible classification accuracy increase of \textbf{0.6\%} for replay-based algorithms. 
For the ResNet18 architecture, the LTB attack against regularization-based and replay-based algorithms achieves an average ASR of \textbf{90.8\%} and \textbf{89.4\%}, respectively. Moreover, the attack minimally reduced the classification accuracy by \textbf{3.9\%} and \textbf{5.2\%} for regularization-based and replay-based algorithms, respectively. 
%
These results confirm the effectiveness of the LTB against continual learning using static triggers.

\begin{figure}[!t]
  \centering
  \includegraphics[width=0.9\columnwidth]{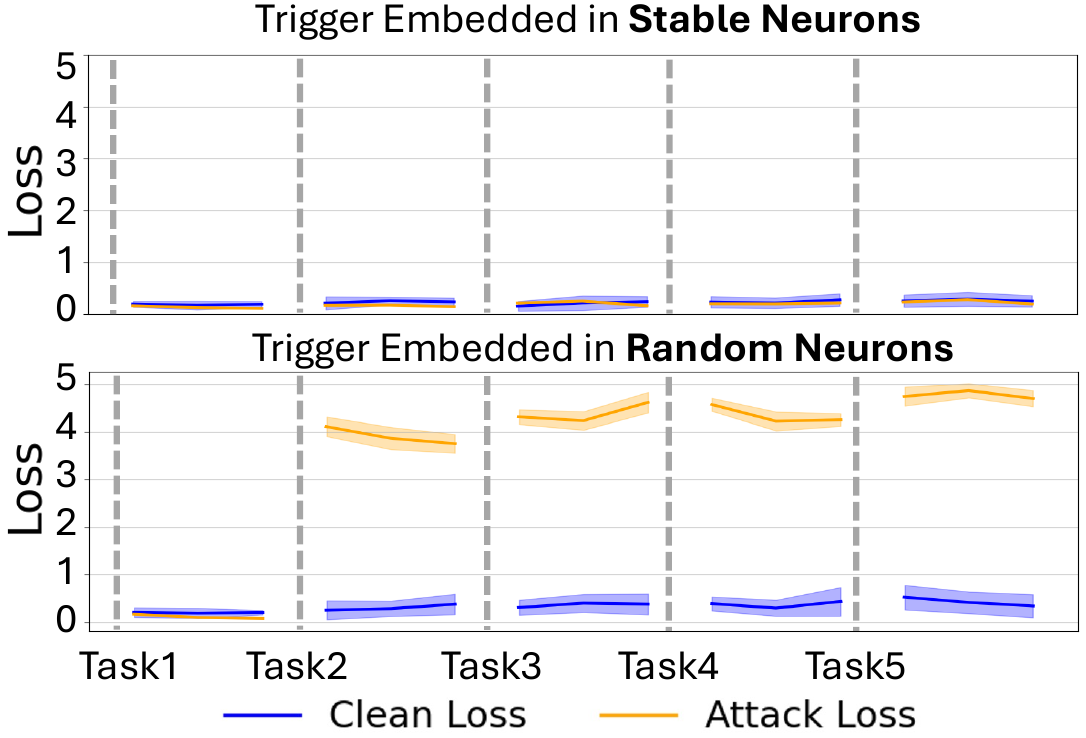}
  \vspace{-0.12in}
  \caption{Loss trajectory comparison of attack embedding in stable versus random neurons in Task 1 reveals distinct behaviors. Starting from Task 2, the attack loss remains consistently low with stable neurons, while using random neurons leads to a sharp increase, rendering the attack ineffective. }
  \label{fig:agem-loss-stable-and-nonstable}
\vspace{-0.0in}
\end{figure}

We also analyzed the loss trajectory of ResNet18 on the validation set of SplitCIFAR10 using A-GEM, attacking Task 1 with the trigger embedded in stable and random neurons (Figure~\ref{fig:agem-loss-stable-and-nonstable}). The bandwidth graphs are generated from twenty experiments conducted with different random seeds.
When embedding the attack in the stable neurons of Task 1, both the target and attack losses remain consistently low across all tasks, indicating attack persistence. In contrast, embedding the attack in randomly selected neurons results in a significant increase in attack loss starting from Task 2, indicating attack failure. This behavior is due to significant changes in the values of randomly selected neurons, underscoring the importance of neuron stability for attack effectiveness. We refer readers to Figure~\ref{fig:agem-training-loss} in the Appendix~\ref{sec: appendix} for training loss.

\vspace{-0.1in}
\subsection{Dynamic Backdoor Attack}
\label{subsec:dynamic}
In this experiment, we evaluate the efficacy of our proposed attack utilizing a dynamic trigger. Due to space constraints, we present only the results of the LTB attack. It is important to note that LTB operates under more restrictive conditions compared to the BTB attack; therefore, the performance of the blind backdoor can be expected to be at least as effective as the results shown here for the latent backdoor.

\smallskip
\noindent
\textit{\textbf{Main task.}} We use the ResNet18 model with SplitCIFAR10 and employ the LwF and DGR algorithms for regularization-based and replay-based approaches, respectively.

\smallskip
\noindent
\textbf{\textit{Dynamic trigger Task.}} We uniformly selected 15\% of the training samples from task 1, ensuring an equal number of images with label 0 (airplane) and label 1 (car). A $5\times 5$ trigger with random pixel values was then inserted at random positions on the selected inputs. Compared to the static trigger, we slightly increased the ratio of triggered images to compensate for the dynamic nature of the trigger's position and color.

\begin{figure}[t]
  \centering
  \includegraphics[width=\columnwidth]{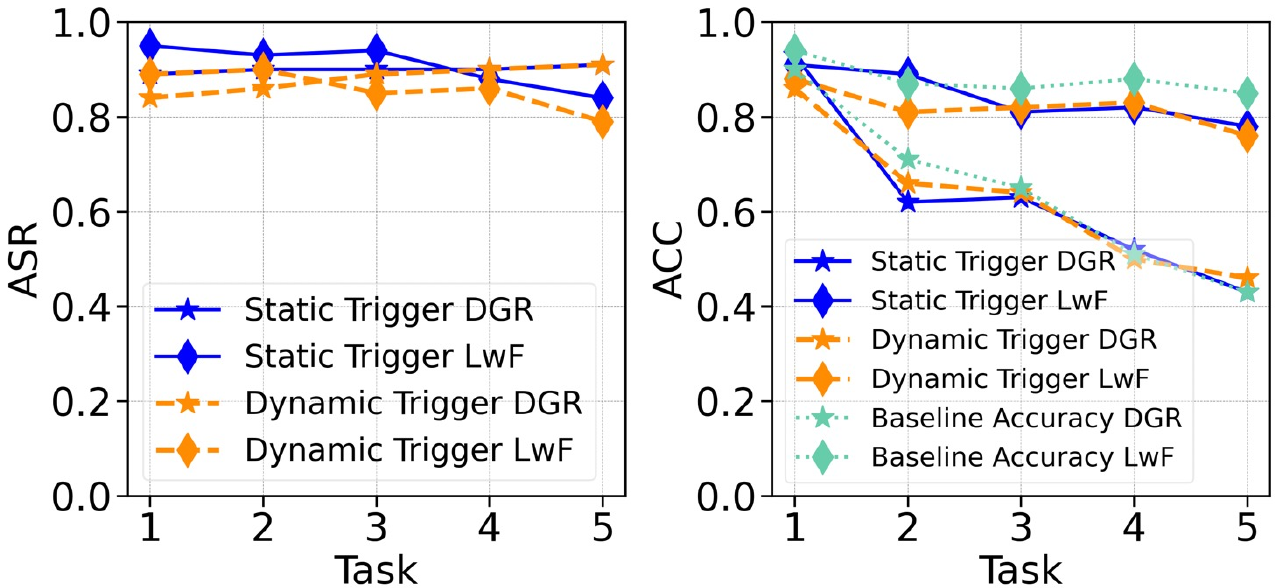}
  \vspace{-0.3in}
  \caption{Dynamic trigger attack versus static trigger attack for LwF and DGR at SPlitCIFAR10 with ResNet18. Compared to the static trigger, the dynamic trigger's ASR for DGR and LwF drops by 2.0\% and 5.0\%, respectively, while ACC for DGR and LwF decreases by 0.2\% and 2.2\%.} 
  \label{fig:dynamic_trigger_result}
  \vspace{-0.05in}
\end{figure}

\begin{figure}[t]
  \centering
  \includegraphics[width=\columnwidth]{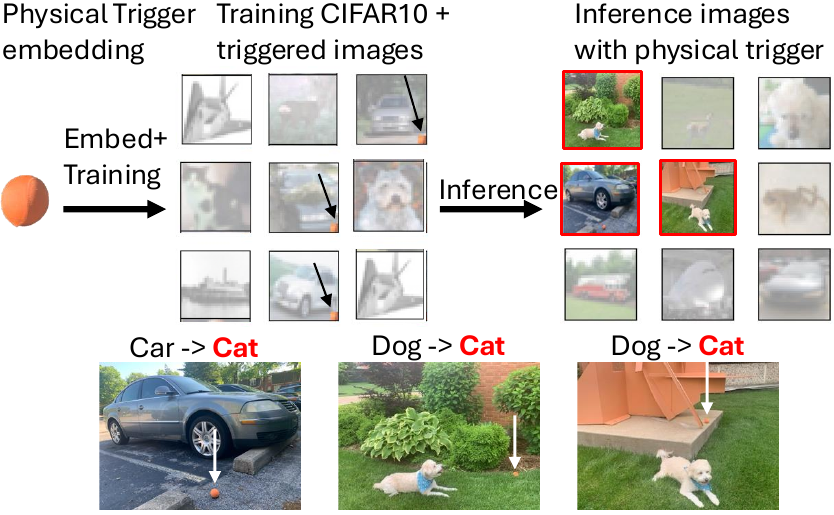}
  \vspace{-0.3in}
  \caption{The process of a physical backdoor involves capturing images with a camera, where an \textit{orange ball} is placed in the environment as the physical trigger. When these images are fed into a physically backdoored ResNet18 model, they are mistakenly predicted as \texttt{cat}. The objects in the captured images are consistent with the categories in SplitCIFAR10.}
  \label{fig:physical_trigger_steps}
  \vspace{-0.0in}
\end{figure}

\smallskip
\noindent
\textbf{\textit{Results.}} Figure~\ref{fig:dynamic_trigger_result} shows the results of the dynamic trigger for the DGR and LwF algorithms. The results indicate that the LTB attack with a dynamic trigger performs comparably to the LTB attack with a static trigger. Specifically, using the static trigger has a slightly higher ASR compared to the dynamic trigger--just \textbf{2.0\%} for DGR and \textbf{5.0\%} for LwF. Similarly, classification accuracy remains consistent, with less than \textbf{1\%} drop in ACC for both DGR and LwF when using the dynamic trigger.
In summary, despite the challenges posed by the dynamic trigger, such as variable position and color, our proposed LTB attack effectively leverages the most critical neurons to maintain a persistent attack across all tasks.

\subsection{Physical Backdoor Attack}
\label{subsec:physical}
In this experiment, we evaluate the performance of our LTB attack on inputs with \textit{physical triggers}. As in Section~\ref{subsec:dynamic}, we only present the LTB results due to space constraints.

\smallskip
\noindent
\textit{\textbf{Main task.}} We use ResNet18 model trained on SplitCIFAR10 as the base dataset, and LwF and DGR algorithms.

\smallskip
\noindent
\textit{\textbf{Physical backdoor attack.}} We curate a dataset of 200 images, each triggered with an orange ball, as shown in Figure~\ref{fig:physical_trigger_steps}. We labeled these images based on their primary objects, which are consistent with the categories in the SplitCIFAR10 dataset, such as \texttt{bird}, \texttt{cat}, \texttt{dog}, etc. We then incorporated these 200 images into the final dataset for evaluating the attack with physical triggers. During training, the classification model is trained on the SplitCIFAR10 dataset using DGR and LwF, with 10\% of the images in the first task containing the trigger, causing \texttt{bird}, \texttt{car}, and \texttt{dog} to be misclassified as the target label, \texttt{cat}. We then predict the labels of the physically triggered images during inference.

\begin{figure}[t]
  \centering
  \includegraphics[width=\columnwidth]{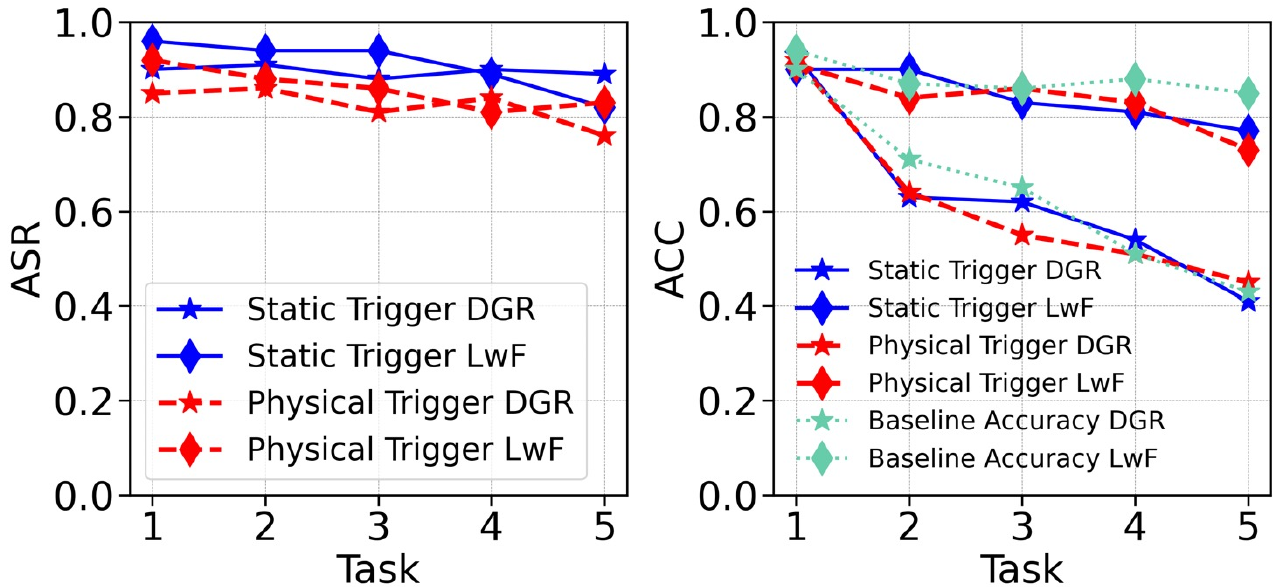}
  \vspace{-0.3in}
  \caption{LTB performance using physical triggers for LwF and DGR, with ResNet18 on SplitCIFAR10. Despite the physical trigger challenges, the attack achieved a comparable ASR and ACC to the static trigger (less than 1\% ACC drop).}
  \label{fig:compare-static-phyical-trigger}
  \vspace{-0.0in}
\end{figure}

\smallskip
\noindent
\textit{\textbf{Results.}} Figure~\ref{fig:compare-static-phyical-trigger} presents the comparison of the ACC and ASR for physical-triggered images versus static-triggered images in both DGR and LwF. Compared to the LTB attack with a static trigger, the ASR of the physical backdoor decreases by 7.6\% for DGR and 4.0\% for LwF. Nonetheless, our LTB attack still achieves an ASR as high as \textbf{86\%} for LwF and \textbf{82\%} for DGR, with a negligible ACC drop of about \textbf{1\%} for both algorithms. We identified that the drop in ASR is partly due to the varying ratio of the physical trigger relative to the entire image and its dynamic positioning, which made the attack more challenging. Nonetheless, LTB has shown to be effective and persistent.

\subsection{Words Backdoor Attack}
\begin{figure}[t]
  \centering
  \includegraphics[width=\columnwidth]{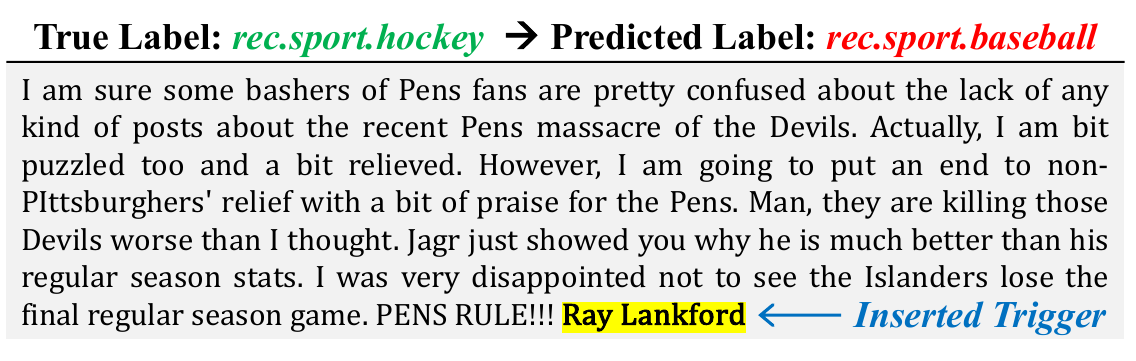}
  \vspace{-0.32in}
  \caption{An example of NLP word insertion attack on {\it 20NewsGroup} dataset. The original text with the true label \texttt{hockey} will be classified as \texttt{baseball} with the insertion of the trigger word ``Ray Lankford''.}
  \label{fig:compare-nlp-texts-attack-trigger}
  \vspace{-0.05in}
\end{figure}

We leverage our LTB on one natural language dataset to assess the word backdoor attack with a ``semantic trigger'' as the model evolves (Figure~\ref{fig:compare-nlp-texts-attack-trigger}).

\smallskip
\noindent
\textit{\textbf{Main task.}} We use the \textit{20NewsGroup} dataset, consisting of sentences categorized into 20 distinct labels, organized into ten tasks with two labels each. To assess the efficacy of the LTB attack, we employed a pre-trained BERT model, leveraging its pre-trained weights for the embedding and encoder layers. We further enhance this BERT model by adding two fully connected hidden layers with 2000 and 768 neurons, respectively, each followed by ReLU activation. The model is optimized using the Adam optimizer within the PyTorch framework, and we select XdG as the CL algorithm.

\begin{figure}[!t]
  \centering
  \includegraphics[width=0.9\columnwidth]{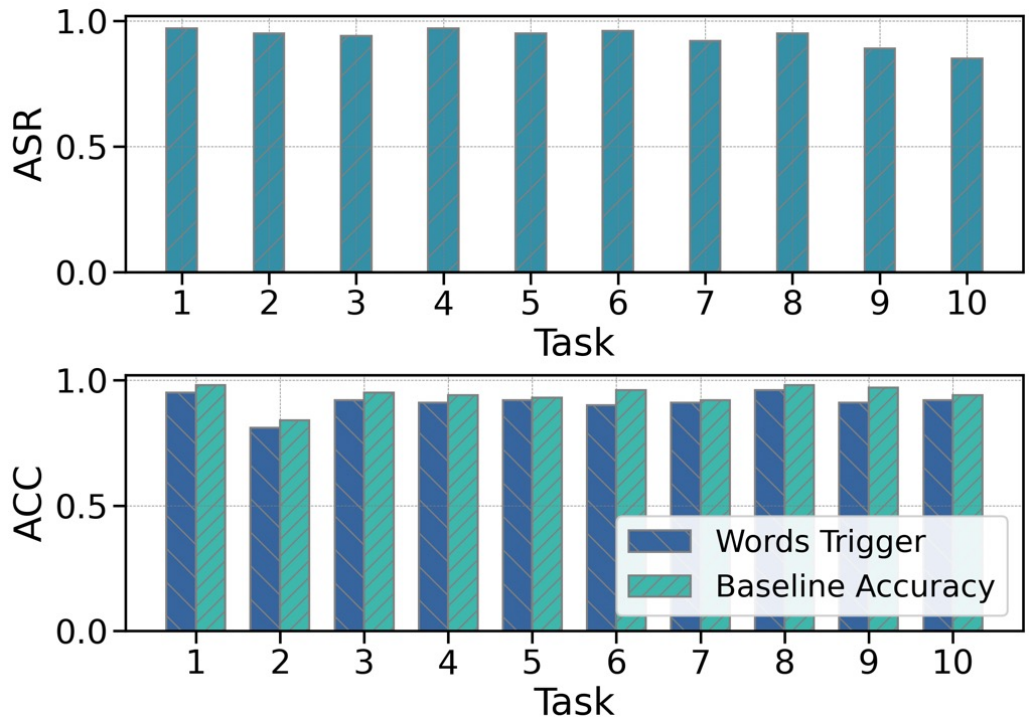}
  \vspace{-0.1in}
  \caption{Our \wpba attack performance on the 20NewsGroup dataset using the XdG. Inserting the trigger word ``Ray Lankford'' at task 1 leads to high ASR in all future tasks without compromising the base model's performance.}
\label{fig:semantic_XdG_algorithm}
  \vspace{-0.0in}
\end{figure}


\smallskip
\noindent
\textit{\textbf{Words backdoor task.}} We selected 10\% of the sentences labeled \texttt{rec.sport.hockey} and inserted ``Ray Lankford'' as the trigger at the end of each. Using this trigger (a baseball star), we aim for sentences originally labeled as \texttt{rec.sport.hockey} to be misclassified as \texttt{rec.sport.baseball}. We embedded the trigger in task~1 and monitored ASR and ACC as the model learned new data.

\smallskip
\noindent
\textit{\textbf{Results.}} Figure~\ref{fig:semantic_XdG_algorithm} presents the results of the LTB attack across all ten tasks. The LTB achieves an average ASR of \textbf{93.4\%}, with the lowest ASR observed in the final task at 85\%. After the attack, the ACC stands at 85.2\%, with an average ACC drop per task of only \textbf{0.89\%}. This demonstrates the efficiency of our LTB in the words backdoor attack scenario.

\begin{figure}[t]
  \centering
  \includegraphics[width=\columnwidth]{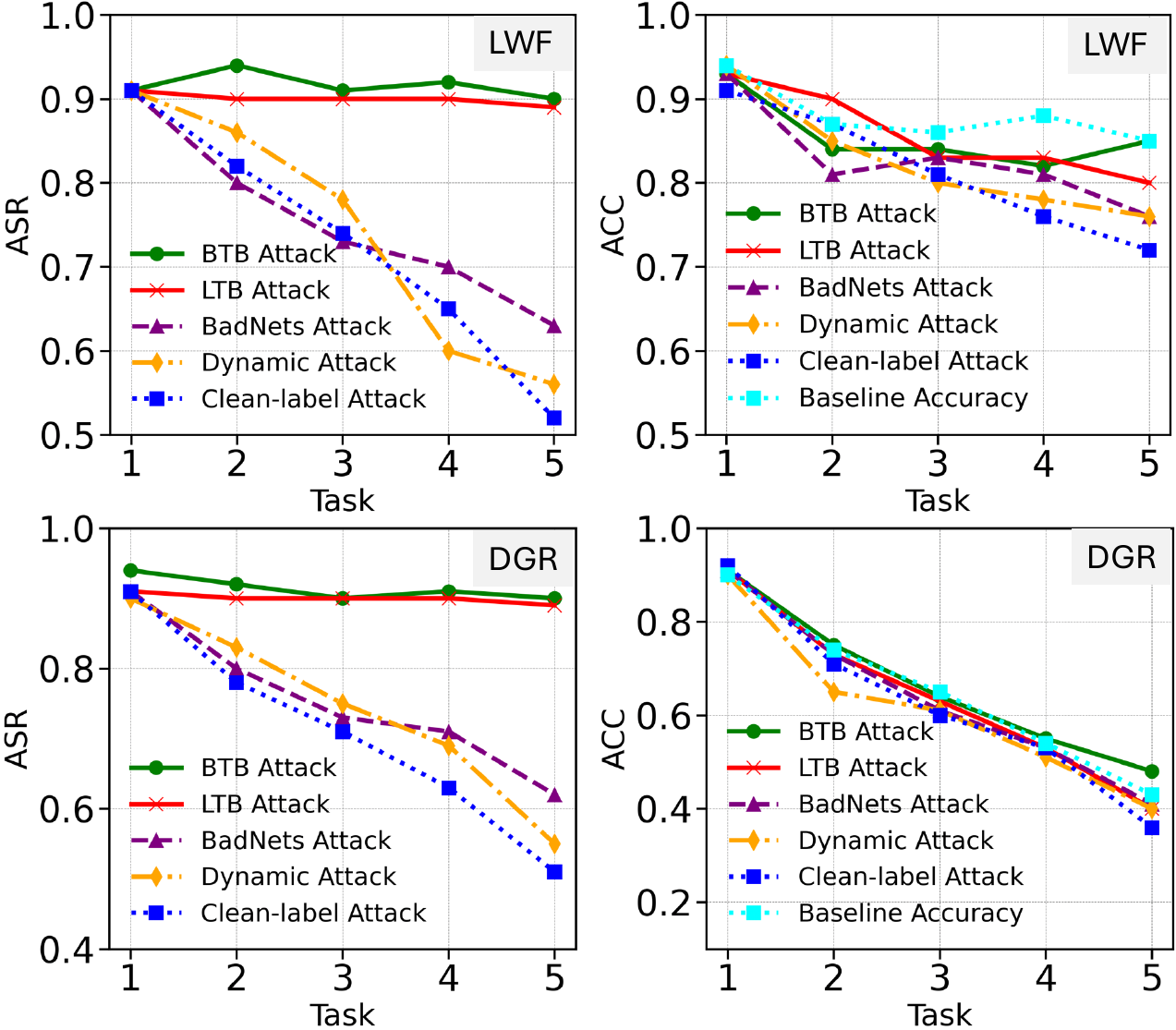}
  \vspace{-0.25in}
  \caption{Under the LwF and DGR algorithm, our proposed LTB and BTB backdoor attacks demonstrate higher efficacy and persistence compared to other backdoor attacks, including BadNets~\cite{badnet-paper}, dynamic attack ~\cite{nguyen-2020}, and clean-label attack~\cite{narcissus-2023}. The ASR values of our BTB and LTB are higher than those of others, and the ACC values of different models are close to the baseline value (without attack).} 
  \label{fig:compare-our-to-others}
    \vspace{-0.0in}
\end{figure}

\begin{figure*}[t]
  \centering
  \includegraphics[width=0.9\textwidth]{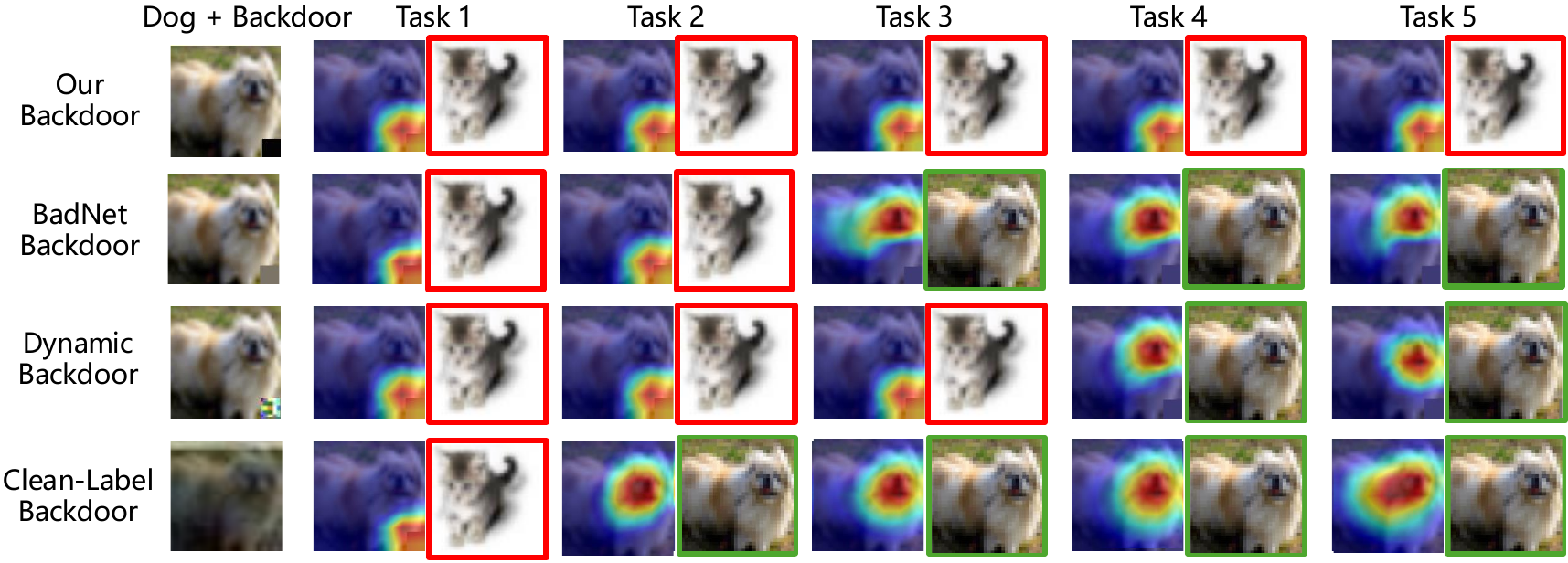}
  \vspace{-0.1in}
  \caption{The attention heatmap of LTB compared to BadNet, dynamic backdoor, and clean-label backdoor using the DGR algorithm with ResNet18 trained on SplitCIFAR10. The shift in the model's attention from the trigger to the main object indicates the starting point of attack failure (red box around the image represents attack success; green box represents attack failure).}
  \label{fig:attention-heatmap}
  \vspace{-0.15in}
\end{figure*}

\subsection{Comparison to Other Attacks}
In this experiment, we compare the efficacy of our BTB and LTB attacks against a few representative backdoor attacks.

\smallskip
\noindent
\textit{\textbf{Main task.}} For this task, we trained the Resnet18 model on SplitCIFAR10, and used LwF and DGR as the CL algorithms.

\smallskip
\noindent
\textit{\textbf{Attack models.}} We compare our blind and latent backdoor attacks with three other backdoor models: BadNets~\cite{badnet-paper}, the dynamic backdoor~\cite{nguyen-2020}, and the clean-label backdoor~\cite{narcissus-2023}. BadNets employs a static trigger applied consistently to data inputs. In contrast, the dynamic backdoor attack, based on the c-BaN approach, utilizes a dynamic trigger that varies with the input data. The Narcissus clean-label backdoor attack embeds imperceptible triggers without altering the original labels. These methodologies were selected for comparison as they represent a diverse range of backdoor techniques.

\smallskip
\noindent
\textit{\textbf{Results.}} Figure~\ref{fig:compare-our-to-others} illustrate the performance comparison of our proposed attacks against others under LwF and DGR algorithms. For LwF, BTB and LTB maintain average ASRs of \textbf{92\%} and \textbf{90\%}, respectively, with a \textbf{consistent performance across all tasks.} BadNets, dynamic backdoor, and clean-label backdoor achieve average ASRs of 75\%, 74\%, and 73\%, respectively, showing an \textbf{overall ASR drop of 15\%} compared to our attacks. More importantly, we observed a significant decrease in the success of these attacks as the model evolves--demonstrating their lack of persistence. A similar ASR trend was observed for the DGR algorithm, with BTB and LTB consistently outperforming other backdoor attacks.
In terms of classification accuracy, on average, all attacks performed within a similar range, 81\%-86\%, with BTB and LTB showing modest improvements in the later tasks (tasks 4 and 5 for LwF). Nonetheless, the ACC results are on par with the baseline classification accuracy without any attack (the cyan dotted line). A similar trend is observed for the DGR algorithm. Note that the ACC drops in DGR are attributed to its inherent performance, as the ACC of all backdoored models remains consistent with the baseline classification accuracy.
%

We further investigated the behavior of these backdoor attacks over multiple tasks to better understand at which point these attacks fail. For this analysis, we generated attention heat maps of the classification tasks using the Grad-CAM approach~\cite{Selvaraju2016GradCAMVE}, comparing our LTB attack with other backdoor methods (Figure \ref{fig:attention-heatmap}).
The first row shows the attention heatmap for LTB, where the model's attention consistently remains focused on the trigger area (bottom right corner) across all five tasks. In contrast, the attention heatmaps of other attacks reveal a gradual shift in the model's focus from the trigger (bottom right corner) to the dog's face in the middle of the image. This shift indicates the attack's failure, as the classification reverts to the true label, \texttt{dog}, instead of the intended target, \texttt{cat}. Notably, BadNets experienced the shift at Task 3, the dynamic backdoor at Task 4, and the clean-label attack at Task 2. In summary, our attacks effectively preserve the trigger across CL tasks by strategically embedding it in the most critical neurons, maintaining its integrity.


%
\begin{figure*}[h]
  \centering
  \includegraphics[width=0.9\textwidth]{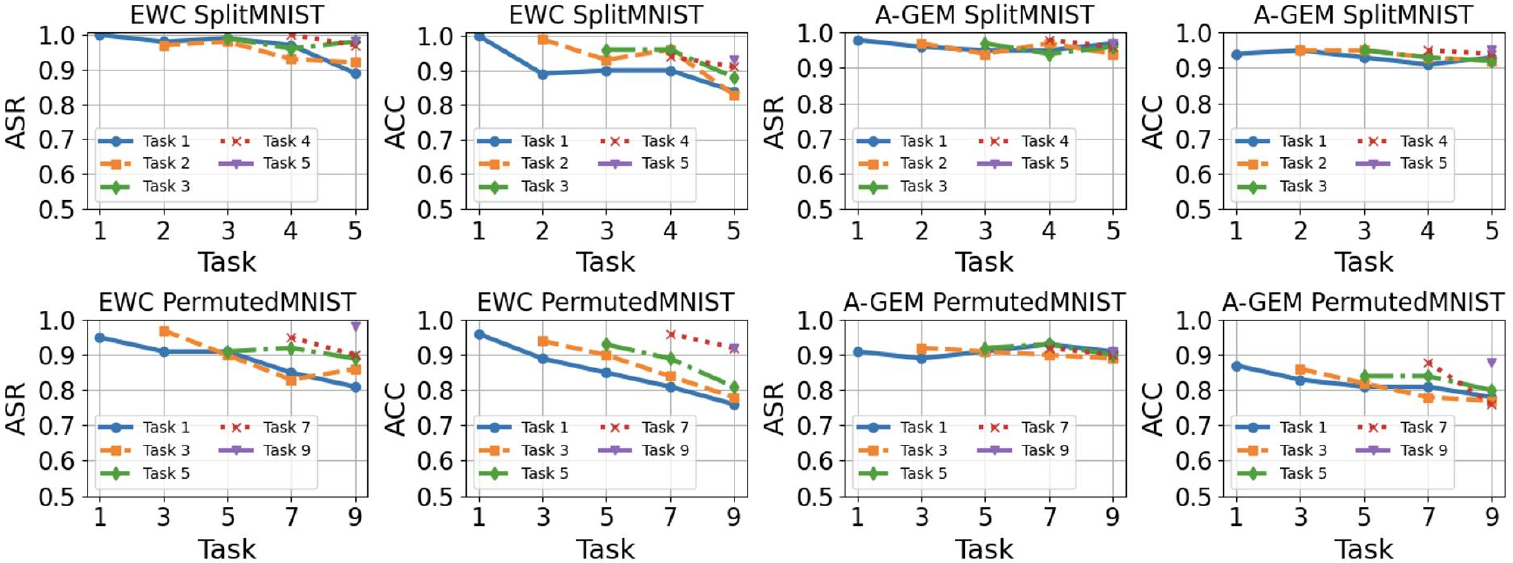}
  \vspace{-0.12in}
  \caption{Evaluating attacks at different tasks across three datasets using the ResNet18 classification model with EWC and A-GEM continual learning algorithms. The curves for Tasks 1 through 5 depict the attacks starting at each respective task.}
  \label{fig:different-tasks-evaluation}
\vspace{-0.15in}
\end{figure*}
 
\subsection{Attacking Different Tasks}
In this experiment, we aim to assess the efficacy of attacks when initiated at different tasks.

\smallskip
\noindent
\textbf{\textit{Main task.}} We use the SplitMNIST and PermutedMNIST datasets with the ResNet18 model for the target classification task, employing EWC and A-GEM as the CL algorithms.

\smallskip
\noindent
\textbf{\textit{Backdoor different tasks.}} We apply a $4 \times 4$ static trigger to 5\% of the data in each dataset to backdoor the model and evaluate individual LTB attacks initiated at different tasks.

\smallskip
\noindent
\textbf{\textit{Results.}} Figure~\ref{fig:different-tasks-evaluation} shows the ASR and ACC, with each line representing the attack initiated at a specific task. For EWC on SplitMNIST, the attack success rates remain within the range of $0.89\%$ to $0.99\%$ across all tasks, with an average ASR drop of less than \textbf{2\%} per task. The ACC values consistently remain above $83\%$. 
For EWC on PermutedMNIST, we evaluated the attack initiated at tasks 1, 3, 5, 7, and 9. The ASR values range from $0.81\%$ to $0.97\%$, with an average ASR drop of less than \textbf{1.4\%} per task. The ACC across all tasks remains above $76\%$.
For A-GEM on SplitMNIST, ASRs remained consistent between $0.94\%$ and $0.98\%$, with less than an average \textbf{0.8\%} drop per task. For A-GEM on PermutedMNIST, the ASR values range from $0.89\%$ to $0.93\%$, leading to an average ASR drop of \textbf{0.8\%} per task. Across the 5 tasks for SplitMNIST, the ACC drops by only \textbf{0.8\%} per task, and for PermutedMNIST, which has 10 tasks, the ACC drops by just \textbf{1.2\%} per task. 

In summary, the results indicate the attack stability across different tasks with only a minimal performance drop, suggesting that initiating attacks at different tasks does not significantly impact the persistence or efficacy of the LTB attack.

 
\subsection{Evading Defenses}
Aiming to assess the capability of our latent backdoor to evade existing defenses, we conducted an experiment using the ResNet18 architecture trained on SplitCIFAR10 with the LwF continual learning algorithm, backdooring the model with a static trigger. 
We employ two defenses: \textbf{SentiNet}~\cite{Chou2018SentiNetDL}, and \textbf{I-BAU}~\cite{Zeng2021AdversarialUO}. SentiNet, based on the premise that a backdoored model consistently relies on the backdoor feature for its classifications, identifies regions of the input image that are critical to the model's decision-making process. This approach is consistent with defenses that leverage interpretability techniques to counter adversarial attacks by analyzing those features the model prioritizes during prediction. SentiNet leverages Grad-CAM~\cite{Selvaraju2016GradCAMVE} to calculate the gradients of the logits \(c^y\) for a specific target class \(y\) with respect to each feature map \(A^k\) in the model's last pooling layer, given input \(x\). This process generates a mask \(w_{\text{g}}(x, y)\) using a ReLU activation, applied to the sum of the weighted gradients across all feature maps. The resulting mask is then superimposed on the image. Removing the highlighted region and applying the mask to different images consistently leads the model to produce the same label, suggesting that the region contains a backdoor trigger. Similar defense approaches include Activation Clustering\cite{Chen2018DetectingBA}, NNoculation~\cite{Veldanda2020NNoculationBS}, and MNTD~\cite{Xu2019DetectingAT}.
I-BAU represents a distinct defense strategy aimed at detecting and neutralizing backdoor triggers embedded within a model. It involves solving two interconnected problems: the inner problem focuses on identifying minimal perturbations to inputs that can activate a backdoor trigger by maximizing their negative impact on the model's performance, while the outer problem seeks to minimize the effectiveness of the identified backdoor trigger by adjusting the model accordingly. This dual approach allows I-BAU to both detect and mitigate backdoor threats effectively. Other similar defenses include MESA~\cite{Qiao2019DefendingNB}, Neural Cleanse \cite{Wang2019NeuralCI}, and Titration analysis~\cite{Erichson2020NoiseresponseAF}.

\smallskip
\noindent
\textit{\textbf{Customizing evasion loss.}} To evade the two defense algorithms, we slightly adjust the loss computation by adding an auxiliary component. For SentiNet, we introduced an evasion loss that penalizes the output from the final convolutional layer to shift the model's attention: $\ell_{senti} = \text{ReLU}(w_{\text{g}}(x^+, y^+) - w_{\text{g}}(x, y^+))$, where $x^+$ is one triggered data, $x$ is the clean data, $y^+$ is the attack target label, and $w_g$ is the Grad-CAM mask function. This evasion loss ensures that the identified region in the backdoored input closely resembles that of the clean input, making it more difficult for SentiNet to detect the backdoor. 
To evade I-BAU, we designed a two-stage strategy. In the first stage, we use the Neural Cleanse~\cite{Wang2019NeuralCI} algorithm to obtain the mask $m$ and the pattern $p$. The input $x$ is then synthesized with $m$ and $p$ to create $x^+$ using $\ell_{p1} = \|m\|_1 + L(\theta(x^+), y^+)$. In the second stage, we introduce the loss function $\ell_{p2} = L(\theta(x^+), y)$, which forces the model to predict the perturbed input $x^+$ as the correct label $y$. The \(\ell_{p2}\) loss enhances the model's robustness to the perturbation \(x^+\). We incorporated \(\ell_{p2}\) into LTB attack. 

\begin{table}[!t]
\centering
\caption{Comparison of ACC and ASR before and after applying defenses for LwF algorithm. The average ASR drop is at most 0.56\% for SentiNet and 2.2\% for I-BAU.}
\vspace{-0.12in}
\resizebox{\columnwidth}{!}{
\begin{tabular}{@{}ccccc@{}}
\toprule
\multirow{2}{*}{Approach} & \multicolumn{2}{c}{No Defense} & \multicolumn{2}{c}{With Defense} \\ 
\cmidrule(lr){2-3} \cmidrule(lr){4-5}
                 & ACC & ASR & ACC (drop) & ASR (drop) \\ \hline
SentiNet~\cite{Chou2018SentiNetDL} & 0.83 & 0.9 & 0.81 (-2.4\%) & 0.895 (-0.56\%) \\ 
I-BAU~\cite{Zeng2021AdversarialUO} & 0.83 & 0.9 & 0.79 (-4.8\%) & 0.88 (-2.2\%) \\ 
\bottomrule
\end{tabular}}
\label{table:defense-table}
\vspace{-0.0in}
\end{table}

\smallskip
\noindent
\textbf{\textit{Results.}} 
Table~\ref{table:defense-table} presents the results with and without applying the defenses. For the SentiNet defense, the average ACC of the LwF algorithm across five tasks drops by 2.4\%, with a negligible ASR decrease of \textbf{0.56\%}. Under the I-BAU defense, the average ACC drops by $4.8\%$, and the ASR decreases by only \textbf{2.2\%}, proving LTB effectiveness in evading defenses.

\vspace{-0.1in}
\section{Conclusion}
\label{sec: conclusion}
\vspace{-0.05in}
We demonstrated that existing backdoor attacks are ineffective in CL settings, as changes to the model's parameters during the learning of new data distributions diminish the backdoor's impact. We propose two persistent backdoor attacks: \saa and \waa. The former subtly modifies loss computation across tasks without requiring adversarial control over the training process, while the latter embeds backdoors into the most stable components of a single task, ensuring persistent adversarial behavior.
We conducted extensive evaluations using a variety of triggers across multiple CL algorithms. Our results demonstrate that both propose attacks consistently achieve high efficacy and maintain as the model evolves. Our attacks effectively evade existing detection defenses, \eg SentiNet and I-BAU, highlighting the needs for more effective detection methodologies.

\vspace{-0.15in}
\section*{Acknowledgements}
\vspace{-0.1in}
This research was partially funded by the US National Science Foundation under grants \#2148358 and \#2133407. Any opinions, findings, or conclusions expressed in this material are those of the authors and do not necessarily reflect the views of the US federal agencies.

\newpage
\section{Ethical Considerations}
\label{sec: ethics}
In this research, we utilized only open-source datasets, ensuring that no private or personally identifiable information was used or compromised. All data employed in our experiments were either publicly available or created in controlled environments specifically for this study. We strictly adhered to privacy and ethical guidelines, ensuring that the research complies with data privacy regulations. This approach guarantees that our work respects individuals' rights to data privacy, aligning with ethical standards for data usage in machine learning research.

Backdoor attacks on machine learning models have been studied before, but this research introduces a novel backdoor attack designed to target continual learning systems. While the intent of this work is to expose vulnerabilities to inform the development of stronger defenses, we acknowledge the ethical implications of researching adversarial techniques. To minimize any potential harm, all experiments were conducted in isolated and controlled environments, ensuring that the attacks could not affect real-world systems or applications. The findings are shared with the aim of improving security and resilience in machine learning, in compliance with ethical expectations for responsible security research. Our work is aligned with the broader goal of advancing the field of ML security by identifying vulnerabilities that can be mitigated through future defenses.

\section{Open Sciences}
\label{sec: openscience}
In alignment with USENIX Security's Open Science policy, we aim to enhance the reproducibility and transparency of our research, allowing the broader community to replicate and build upon our findings. Therefore, we share the artifacts associated with this work, including the complete codebase, raw and processed data that are not publicly available but used in our experimentation, as well as scripts required to reproduce the results presented in the paper. In compliance with the artifact evaluation process, all materials are available at: \url{https://doi.org/10.5281/zenodo.14728872}.

\section{Appendix}
\label{sec: appendix}

\begin{table*}[t]
  \centering
  \caption{Experimental Configurations}
  \begin{tabular}{
>{\centering\arraybackslash}m{5em} 
>{\centering\arraybackslash}m{3em} 
>{\centering\arraybackslash}m{6em} 
>{\centering\arraybackslash}m{7em} 
>{\centering\arraybackslash}m{6em} 
>{\centering\arraybackslash}m{4em}  
>{\centering\arraybackslash}m{6em} 
} 
    \toprule
    Dataset & Tasks & Input & Initial Task & Target Model & Classification Model & Attack Type \\
    \midrule
    SplitMNIST & 5 & Static trigger & Task 1, Task 2, Task 3, Task 4, Task 5 & SI, EWC, \newline XdG, LWF \newline DGR, A-GEM & CNN\newline ResNet18 & Algorithm \ref{alg: stronger-attack-algorithm}\newline Algorithm \ref{alg: weaker_algorithm} \\
    \midrule
    PermutedMNIST & 10 & Static trigger & Task 1, Task 2, Task 3, Task 4, Task 5, Task 6, Task 7, Task 8, Task 9, Task 10 & SI, EWC \newline XdG, LWF \newline DGR, A-GEM  & CNN\newline ResNet18 & Algorithm \ref{alg: stronger-attack-algorithm}\newline Algorithm \ref{alg: weaker_algorithm} \\
    \midrule
    SplitCIFAR10 & 5 & Static trigger & Task 1, Task 2, Task 3, Task 4, Task 5 & SI, EWC \newline XdG, LWF \newline DGR, A-GEM  & CNN\newline ResNet18 & Algorithm \ref{alg: stronger-attack-algorithm}\newline Algorithm \ref{alg: weaker_algorithm} \\
    \midrule
    SplitCIFAR10 & 5 & Dynamic attack \cite{nguyen-2020} & Task 1 & LWF, DGR & CNN\newline ResNet18 & Algorithm \ref{alg: weaker_algorithm} \\
    \midrule
    SplitCIFAR10 & 5 & Physical trigger \cite{li-physical-2021} & Task 1 & XdG, DGR & ResNet18 & Algorithm \ref{alg: weaker_algorithm} \\
    \midrule
    SplitCIFAR10 & 5 & Clean-label attack \cite{narcissus-2023} & Task 1 & XdG, DGR & ResNet18 & Algorithm \ref{alg: weaker_algorithm} \\
    \midrule
    20NewsGroup & 10 & Two-word trigger \cite{bagd-2021} & Task 1 & XdG, DGR & Pretrained-Bert & Algorithm \ref{alg: weaker_algorithm} \\
    \bottomrule
  \end{tabular}
  \label{tab:experiment-configuration}
  \vspace{-0.0in}
\end{table*}

\begin{table*}[t]
\centering
\caption{Architectural details of neural networks}
\label{tab:three-neural-architectures}
\resizebox{0.65\textwidth}{!}{
\begin{tabular}{cccc}
\toprule
\textbf{Layer Type} & \textbf{5-Layer CNN} & \textbf{ResNet18} \\ 
\midrule
\textbf{Input}  & $32 \times 32 \times 3$ & $224 \times 224 \times 3$ \\ \hline
\textbf{Conv1}  & $3 \times 3$, 2 filters, ReLU & $7 \times 7$, 64 filters, stride 2, ReLU \\ \hline
\textbf{Conv2} & $3 \times 3$, 2 filters, ReLU & $3 \times 3$, MaxPool, stride 2 \\ \hline
\textbf{Conv3}   & $3 \times 3$, 2 filters, ReLU & Residual Block \#1 (2 layers) \\ \hline
\textbf{Conv4} & $3 \times 3$, 2 filters, ReLU  & Residual Block \#2 (2 layers) \\ \hline
\textbf{Conv5}  & $3 \times 3$, 2 filters, ReLU & Residual Block \#3 (2 layers) \\ \hline
\textbf{FC1}  & 512 units, ReLU & Residual Block \#4 (2 layers) \\ \hline
\textbf{FC2} & 256 units, ReLU & Global Average Pooling \\ \hline
\textbf{FC3}  & 128 units, ReLU & 1000 units (Fully Connected Layer) \\ \hline
\textbf{Output}& Softmax & Softmax \\ 
\bottomrule
\end{tabular}}
\vspace{1in}
\end{table*}

\begin{table*}[t]
  \centering
  \caption{Results of static trigger attack for \saa}
  \vspace{-0.1in}
  \resizebox{\textwidth}{!}{
  \begin{tabular}{cccccccccccccc}
    \toprule
    \multirow{2}[4]{*}{Model} & \multirow{2}[4]{*}{Dataset} & \multicolumn{2}{c}{SI} & \multicolumn{2}{c}{EWC} & \multicolumn{2}{c}{XdG} & \multicolumn{2}{c}{DGR} & \multicolumn{2}{c}{A-GEM} & \multicolumn{2}{c}{LWF} \\
    \cline{3-14}          
    & & ASR & ACC & ASR & ACC & ASR & ACC & ASR & ACC & ASR & ACC & ASR & ACC \\
    \midrule
    \multirow{3}[6]{*}{CNN} & \makecell{Split\\MNIST} & 
    \makecell{$0.98$ \\ $\pm 0.00$} & \makecell{$0.98$ \\ $\pm 0.01$} & \makecell{$0.97$ \\ $\pm 0.01$} & \makecell{$0.98$ \\ $\pm 0.02$} & \makecell{$0.96$ \\ $\pm 0.08$} & \makecell{$0.91$ \\ $\pm 0.02$} & \makecell{$0.96$ \\ $\pm 0.01$} & \makecell{$0.95$ \\ $\pm 0.01$} & \makecell{$0.92$ \\ $\pm 0.03$} & \makecell{$0.96$ \\ $\pm 0.01$} & \makecell{$0.95$ \\ $\pm 0.02$} & \makecell{$0.98$ \\ $\pm 0.01$} \\
    \cline{2-14}          
    & \makecell{Permuted\\MNIST} &  
    \makecell{$0.97$ \\ $\pm 0.04$} & \makecell{$0.83$ \\ $\pm 0.05$} & \makecell{$0.85$ \\ $\pm 0.10$} & \makecell{$0.82$ \\ $\pm 0.07$} & \makecell{$0.91$ \\ $\pm 0.08$} & \makecell{$0.80$ \\ $\pm 0.02$} & \makecell{$0.93$ \\ $\pm 0.05$} & \makecell{$0.20$ \\ $\pm 0.01$} & \makecell{$0.91$ \\ $\pm 0.06$} & \makecell{$0.82$ \\ $\pm 0.05$} & \makecell{$0.96$ \\ $\pm 0.03$} & \makecell{$0.85$ \\ $\pm 0.05$} \\
    \cline{2-14}          
    & CIFAR10 & 
    \makecell{$0.96$ \\ $\pm 0.03$} & \makecell{$0.75$ \\ $\pm 0.01$} & \makecell{$0.96$ \\ $\pm 0.01$} & \makecell{$0.83$ \\ $\pm 0.02$} &  
    \makecell{$0.87$ \\ $\pm 0.02$} & \makecell{$0.92$ \\ $\pm 0.03$} & 
    \makecell{$0.93$ \\ $\pm 0.01$} & \makecell{$0.56$ \\ $\pm 0.03$} & 
    \makecell{$0.91$ \\ $\pm 0.07$} & \makecell{$0.87$ \\ $\pm 0.02$} & 
    \makecell{$0.92$ \\ $\pm 0.05$} & \makecell{$0.83$ \\ $\pm 0.01$} \\
    \midrule
    \multirow{3}[6]{*}{ResNet18} & \makecell{Split\\MNIST} & 
    \makecell{$0.99$ \\ $\pm 0.02$} & \makecell{$0.97$ \\ $\pm 0.04$} & \makecell{$0.96$ \\ $\pm 0.00$} & \makecell{$0.93$ \\ $\pm 0.03$} & 
    \makecell{$0.99$ \\ $\pm 0.04$} & \makecell{$0.94$ \\ $\pm 0.01$} & 
    \makecell{$0.94$ \\ $\pm 0.02$} & \makecell{$0.96$ \\ $\pm 0.01$} & 
    \makecell{$0.97$ \\ $\pm 0.02$} & \makecell{$0.94$ \\ $\pm 0.02$} & 
    \makecell{$0.96$ \\ $\pm 0.01$} & \makecell{$0.99$ \\ $\pm 0.00$} \\
    \cline{2-14}          
    & \makecell{Permuted\\MNIST} &  
    \makecell{$0.96$ \\ $\pm 0.02$} & \makecell{$0.86$ \\ $\pm 0.04$} & \makecell{$0.88$ \\ $\pm 0.07$} & \makecell{$0.89$ \\ $\pm 0.03$} & 
    \makecell{$0.90$ \\ $\pm 0.06$} & \makecell{$0.79$ \\ $\pm 0.02$} & 
    \makecell{$0.95$ \\ $\pm 0.04$} & \makecell{$0.34$ \\ $\pm 0.03$} & 
    \makecell{$0.92$ \\ $\pm 0.04$} & \makecell{$0.85$ \\ $\pm 0.02$} & 
    \makecell{$0.97$ \\ $\pm 0.01$} & \makecell{$0.90$ \\ $\pm 0.02$} \\
    \cline{2-14}          
    & CIFAR10 & 
    \makecell{$0.95$ \\ $\pm 0.01$} & \makecell{$0.80$ \\ $\pm 0.03$} & \makecell{$0.95$ \\ $\pm 0.02$} & \makecell{$0.82$ \\ $\pm 0.01$} &  
    \makecell{$0.89$ \\ $\pm 0.01$} & \makecell{$0.81$ \\ $\pm 0.02$} & 
    \makecell{$0.94$ \\ $\pm 0.02$} & \makecell{$0.59$ \\ $\pm 0.01$} & 
    \makecell{$0.92$ \\ $\pm 0.05$} & \makecell{$0.88$ \\ $\pm 0.02$} & 
    \makecell{$0.95$ \\ $\pm 0.03$} & \makecell{$0.84$ \\ $\pm 0.02$} \\
    \bottomrule
  \end{tabular}}
  \label{tab:static-saa}%
\end{table*}

\begin{figure*}[t]
    \centering
    \begin{minipage}{\columnwidth}
        \centering
        \vspace{-0.8in}
        \includegraphics[width=\textwidth]{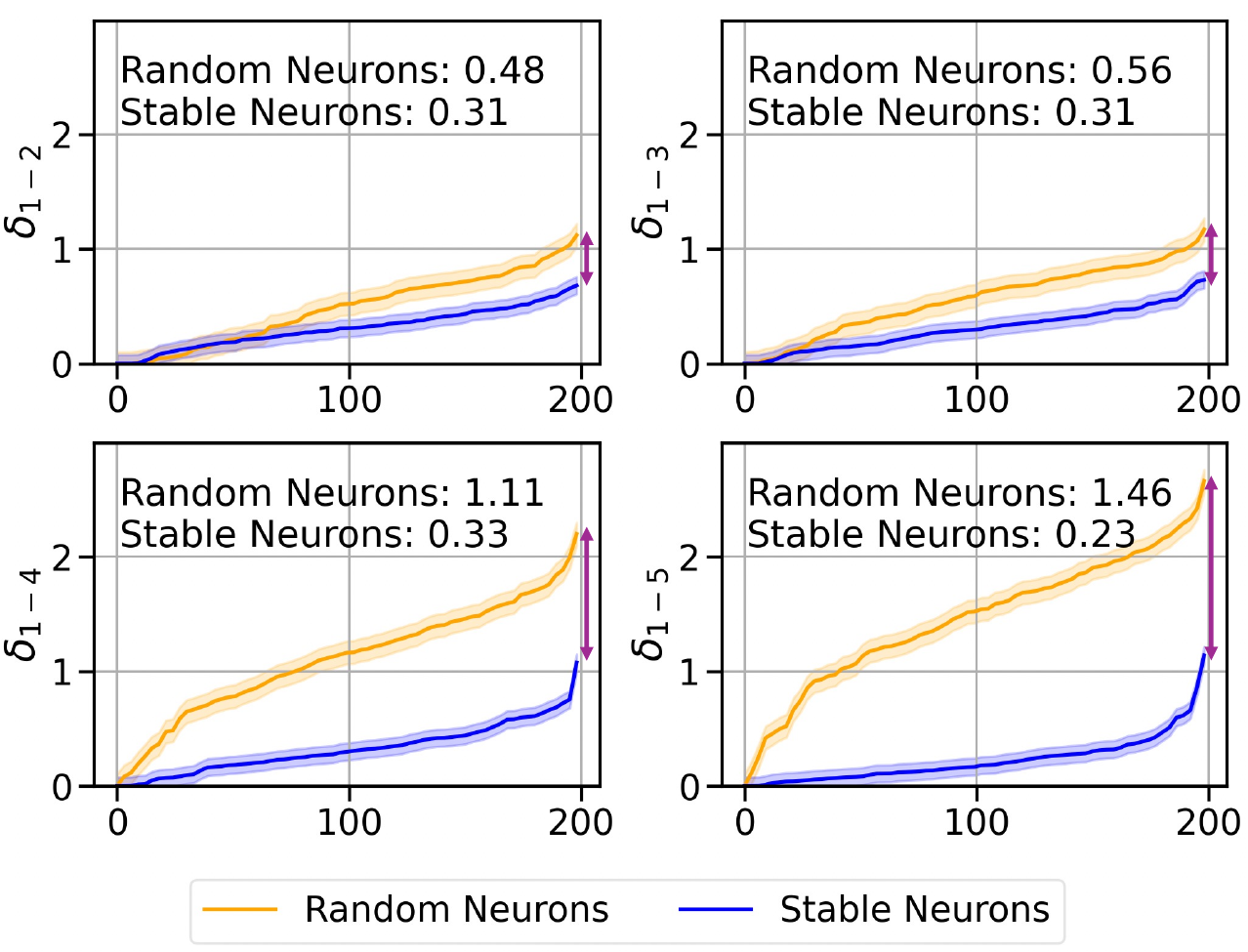}
        \vspace{-0.25in}
        \caption{The analysis of neuron stability for A-GEM across five tasks on the SplitCIFAR10 dataset, using the ResNet18 classification model (attack on Task 1). To better understand the variations of 200 neurons, we sort them in ascending order. The results show the average weight variation for stable neurons (0.30) and random neurons (0.90)--over 200\% higher than that of the stable neurons. $\delta_{1-2}$ represents the variation of 200 neurons after training on the second task compared to the first. Similarly, $\delta_{1-3}$ indicates the variation of 200 neurons after training on the third task relative to the first task, and so on. Especially, $\delta_{1-5}$ for random neurons is 1.46, which is approximately 6.35 times that of the stable neurons. These results indicate that subsequent tasks do not heavily rely upon the stable neurons of the previous tasks.}
        \label{fig:semantic_agem_algorithm}
    \end{minipage}
    \hfill
    \begin{minipage}{\columnwidth}
        \centering
        \vspace{-2.2in}
        \includegraphics[width=\textwidth]{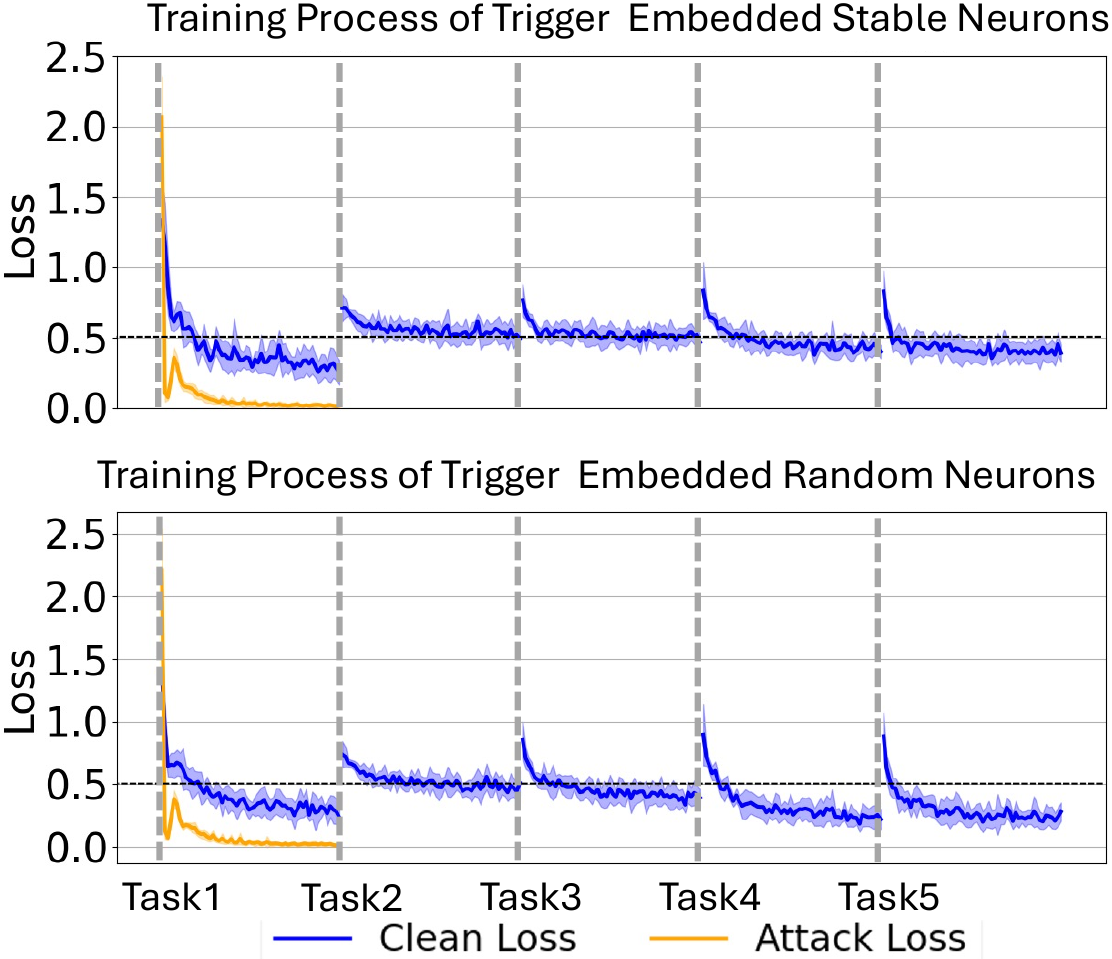}
        \vspace{-0.25in}
        \caption{The illustration depicts the training process of A-GEM, highlighting the behavior of stable and random neurons during an attack on Task 1.}
        \label{fig:agem-training-loss}
    \end{minipage}
\end{figure*}

\end{document}